\documentclass[11pt,letterpaper]{article}
\usepackage[T1]{fontenc}
\usepackage[utf8]{inputenc}
\usepackage[margin=1in]{geometry}
\usepackage{times}
\usepackage[authoryear,round]{natbib}
\setcitestyle{citesep={;},aysep={,},yysep={;}}

\usepackage{amsmath,amsfonts,bm}

\def\eqref#1{equation~\ref{#1}}

\def\1{\bm{1}}

\DeclareMathAlphabet{\mathsfit}{\encodingdefault}{\sfdefault}{m}{sl}
\SetMathAlphabet{\mathsfit}{bold}{\encodingdefault}{\sfdefault}{bx}{n}

\usepackage{amsmath,amssymb,booktabs,graphicx,tabularx}
\usepackage[table]{xcolor}
\usepackage{placeins}
\usepackage{float}
\usepackage{hyperref,url}

\definecolor{fengruiGreen}{RGB}{0,128,0}

\hypersetup{hidelinks,
pdftitle={A Unified Arena for LLM Agents across Diverse Eurogames},
pdfauthor={Fengrui Zhang, Xiang Li, Ruomeng Ding, Jiayi Yuan, Yuchen Huang, Zhaoyang Han, Xiaoliang Wang, Daoyuan Chen, Yaliang Li},
pdfsubject={A unified arena for evaluating LLM agents across ten Eurogames},
pdfkeywords={DECE, LLM agents, Eurogames, competitive evaluation, long-horizon environments}}

\title{A Unified Arena for LLM Agents\\across Diverse Eurogames}
\author{%
  \begin{minipage}{\dimexpr\textwidth-2\tabcolsep\relax}
  \centering\normalsize
  Fengrui Zhang$^{1,2}$,\quad
  Xiang Li$^{3}$,\quad
  Ruomeng Ding$^{4}$,\quad
  Jiayi Yuan,\quad
  Yuchen Huang$^{5}$\\[4pt]
  Zhaoyang Han$^{6}$,\quad
  Xiaoliang Wang$^{2}$,\quad
  Daoyuan Chen$^{1}$,\quad
  Yaliang Li$^{1}$\\[9pt]
  \small
  $^{1}$Alibaba Group\quad
  $^{2}$Nanjing University\\[2pt]
  $^{3}$National University of Singapore\\[2pt]
  $^{4}$Department of Computer Science,
  University of North Carolina at Chapel Hill\\[2pt]
  $^{5}$The Hong Kong University of Science and Technology\\[2pt]
  $^{6}$Peking University
  \end{minipage}%
}
\date{}
\begin{document}
\maketitle
\suppressfloats[t]

\begin{abstract}
Interactive environments offer a practical way to evaluate large language
model (LLM) agents, and competitive games extend this evaluation to decisions
shaped by competing goals and other agents' choices.
Yet existing evaluations confined to individual games or execution frameworks leave unclear
which competitive advantages persist across settings.
We introduce \textbf{DECE} (\textbf{D}iverse \textbf{E}urogame
\textbf{C}ompetitive \textbf{E}nvironments), a unified arena spanning ten
Eurogames with shared interfaces and evaluation protocols.
These games combine lasting strategic commitments, competition over shared
resources, and multiple routes to victory.
Our primary study evaluates 12 model–harness configurations, spanning nine models and four execution harnesses, in 480 three-player matches across ten games. The main and supplementary studies comprise 1,032 completed matches in total.
These comparisons reveal broad cross-game
ranking agreement alongside game-specific strengths.
Matched studies show that the same model can perform differently across
execution harnesses, adding another source of variation to these profiles.
These performance advantages do not consistently translate into better token
or cost efficiency, making resource use an additional consideration when
comparing configurations.
Together, the findings support evaluating complete agent configurations by both
their competitive performance across games and the resources needed to achieve it.

\end{abstract}

\section{Introduction}
\label{sec:introduction}

Large language model (LLM) agents increasingly act through tools in external
environments \citep{yao2023react}. Interactive benchmarks evaluate their
decisions over extended trajectories \citep{liu2024agentbench,zhou2024webarena}.
In competition, opponents pursue their own objectives and change an agent's
available opportunities.
An action's value depends on both its future consequences and opponents'
choices. We ask:
\begin{center}
\emph{How can competitive environments support reliable evaluation of}\\
\emph{complete agent systems across diverse strategic demands?}
\end{center}

Strategy board games make this question tractable. They combine explicit
rules, programmatically executable states and actions, and independently
verifiable outcomes \citep{gaina2020tag,balla2024pytag}.
Their interactions can be represented as
partially observable stochastic games (POSGs) \citep{hansen2004posg}.
Agents act on their observation histories; their actions determine state
transitions and outcomes.
We focus on \emph{Eurogames} (German-style strategy board games), which combine
persistent investments, competition over shared opportunities, and multiple
routes to victory. In \emph{Splendor}, development cards reduce later purchase
costs; \emph{Azul} placements constrain future spatial scoring
\citep{rules_splendor,rules_azul}. These commitments shape both a player's
future options and opponents' opportunities.

Existing multi-game benchmarks and harness studies provide foundations for
competitive evaluation \citep{costarelli2024gamebench,hua2026ptcg}.
Turning these games into a common benchmark presents three challenges.
Agents need consistent observation--action interfaces across diverse rules
and information structures, while retaining their tools and context management.
Analysis must distinguish recurring competitive strength from game-specific
advantages and account for execution settings and resource use.
Reliable comparisons require consistent rule enforcement, a common
\mbox{participant roster}, seat-balanced protocols, and verifiable outcomes.
Matched scenarios and uncertainty estimates support comparisons of execution
settings.

\begingroup
\setlength{\intextsep}{4pt}
\begin{figure}[H]
\centering
\includegraphics[width=\linewidth]{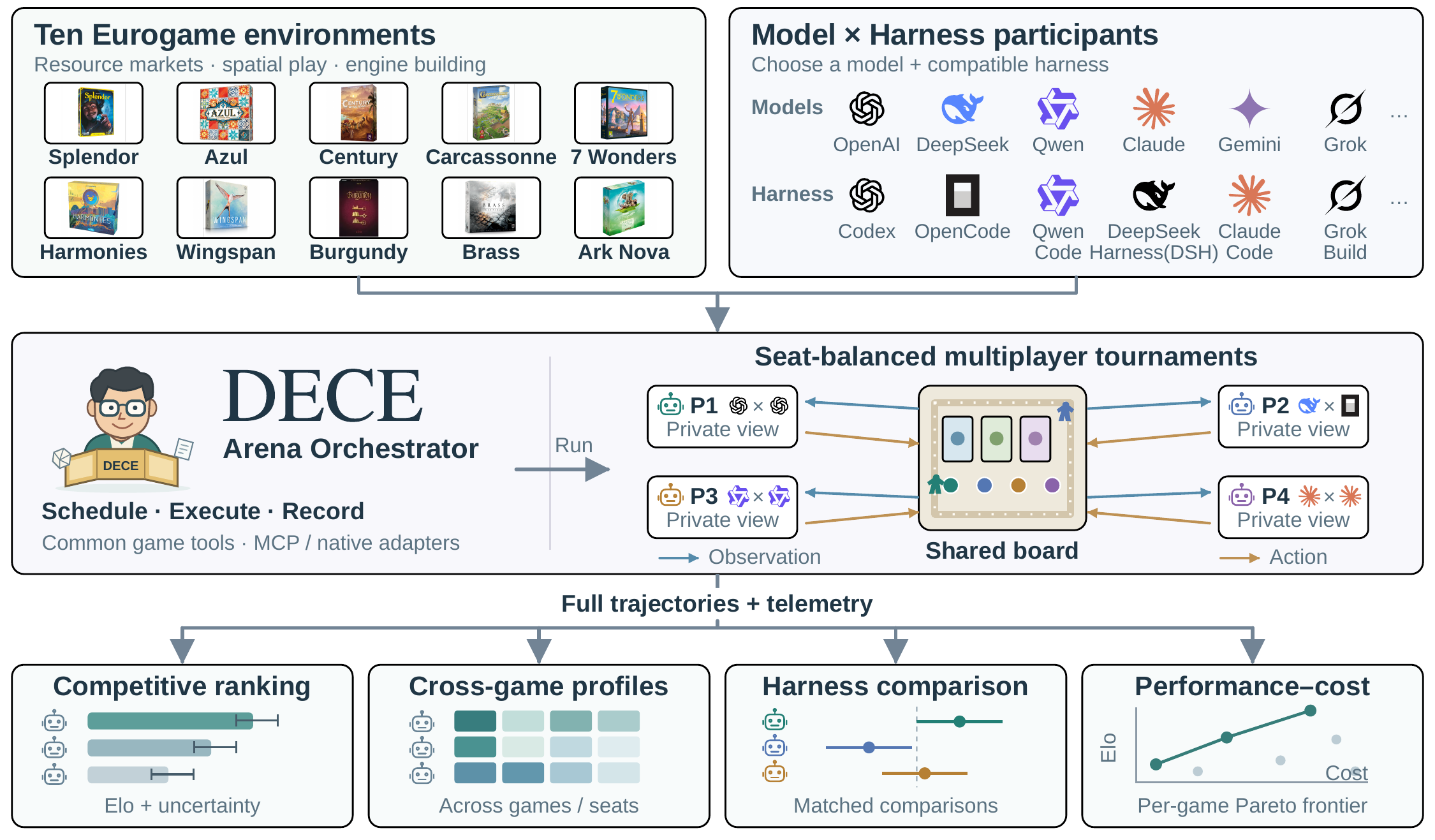}
\caption{\textbf{DECE overview.} Shared interfaces connect games and
model--harness systems. Seat-balanced matches produce trajectories and telemetry
for four analyses. Icons show adapter families; Table~\ref{tab:performance-compact}
lists evaluated configurations. The mascot, pairings, and plots are schematic.}
\label{fig:arena-overview}
\end{figure}
\endgroup

We introduce \textbf{DECE} (\textbf{D}iverse \textbf{E}urogame
\textbf{C}ompetitive \textbf{E}nvironments), a unified evaluation framework
spanning ten executable Eurogames.\footnote{The name also echoes
\emph{d\'{e} c\`{e}}, a Chinese shorthand for Euro-style strategy board
games.} The games vary in their demands for economic development, spatial
commitment, and coordination across scoring routes while retaining a shared
competitive structure. As illustrated in Figure~\ref{fig:arena-overview}, a
common interface exposes seat-specific observations and legal actions to
pluggable execution harnesses. Each participant is a complete
\emph{model--harness configuration}, including its tools and context
management. A common roster and seat-balanced protocols support cross-game
comparisons. Separate matched studies compare the same model across harnesses,
fixing setup seeds, focal seats, and opponent configurations within each scenario.
Game engines enforce rules and scoring; execution records support auditing
and outcome checks. We quantify uncertainty by resampling whole matches or
matched scenarios. We report token use and API-equivalent cost under a fixed
pricing schedule alongside competitive outcomes.

Our evaluation comprises 1{,}032 completed matches across the main and
supplementary studies. The primary study evaluates 12 model--harness
configurations, spanning nine models and four harnesses, in 480 three-player
matches across ten games. Rankings formed on nine games agree with the held-out
game at mean Kendall $\tau_b=0.664$ and 83.9\% pair-order accuracy.
Game-level profiles reveal distinct strengths and local reversals that the
aggregate ranking obscures. Separate matched studies show that the same
model's performance can differ across execution harnesses.
Performance--resource trade-offs vary across games, and token-based and
cost-based comparisons favor different configurations.
These findings support evaluating complete agent systems across multiple
competitive environments, with game-level profiles and resource use reported
alongside aggregate rankings.

Section~\ref{sec:related-work} reviews related benchmarks.
Section~\ref{sec:dece} presents DECE's environments and evaluation protocol.
Section~\ref{sec:experiments} examines cross-game performance profiles
(\textbf{RQ1}), matched harness comparisons (\textbf{RQ2}), and game-specific
performance--resource trade-offs (\textbf{RQ3}).
Section~\ref{sec:discussion} discusses implications for competitive agent
evaluation.

\section{Related Work}
\label{sec:related-work}
Games provide a natural setting for evaluating LLM agents because they combine explicit rules, sequential interaction, and verifiable outcomes. Recent benchmarks span both digital and tabletop environments: Orak, BALROG, and LMGame-Bench evaluate agents across video games
\citep{park2026orak,paglieri2025balrog,hu2026lmgame}
, while GameBench and MindGames study competitive and social reasoning in board, card, and strategy games \citep{costarelli2024gamebench,wang2026mindgames}.
FinBoardBench studies financial decisions and PTCG-Bench examines
card play, harness ablations, self-evolution, and cost
\citep{hu2026finboardbench,hua2026ptcg}.
Complementary work such as TAG and From Code to Play provides executable tabletop environments or evaluates components used by search-based players \citep{gaina2020tag,eberhardinger2024codetoplay}, while LudoBench studies reasoning from fixed game states \citep{peper2026ludobench}.

\paragraph{Position of DECE.} %

DECE compares complete agent configurations across ten Eurogames with
persistent investments, spatial commitments, and multiple scoring routes.
A common roster measures cross-game agreement and local strengths; matched
studies assess harness contrasts, and game-specific frontiers relate
performance to dollars and tokens. Table~\ref{tab:rq-coverage} compares the
benchmark scope.

\begin{table}[!htbp]
\centering\small
\setlength{\tabcolsep}{3pt}
\renewcommand{\arraystretch}{1.08}
\caption{Game scope, decision demands, and evaluation focus of related
tabletop benchmarks. Entries summarize reported tasks and selected analyses.
Appendix~\ref{app:horizon-reporting} gives sources and scope.}
\label{tab:rq-coverage}
\begin{tabularx}{\linewidth}{@{}>{\raggedright\arraybackslash}p{.16\linewidth}
>{\raggedright\arraybackslash}p{.19\linewidth}
>{\raggedright\arraybackslash}p{.14\linewidth}
>{\raggedright\arraybackslash}p{.24\linewidth}
>{\raggedright\arraybackslash}X@{}}
\toprule
Benchmark & Game scope & Interaction & Main decision demands & Evaluation focus\\
\midrule
GameBench \citeyearpar{costarelli2024gamebench}
& 9 board, card and social games
& Players or teams
& Spatial tactics, deck building and social inference
& Competitive play; scaffold comparisons\\
\addlinespace[3pt]
MindGames \citeyearpar{wang2026mindgames}
& 4 strategic and social games
& 2--6 players; rivals or teams
& Opponent inference, signaling, trust and deception
& Social reasoning; arena validity\\
\addlinespace[3pt]
FinBoardBench \citeyearpar{hu2026finboardbench}
& Cashflow, Acquire, Monopoly
& 4-player competition
& Cash flow, investment, mergers and negotiation
& Financial decision analysis\\
\addlinespace[3pt]
PTCG-Bench \citeyearpar{hua2026ptcg}
& Pok\'emon TCG; 5 decks
& 2-player competition
& Card combinations, resource recovery and action order
& Self-evolution; harness and cost analysis\\
\addlinespace[3pt]
From Code to Play \citeyearpar{eberhardinger2024codetoplay}
& 12 TAG tabletop games
& 2--4-player tournaments
& Deck building, drafting and tactical card play
& Generate evaluators for search agents\\
\addlinespace[3pt]
LudoBench \citeyearpar{peper2026ludobench}
& 5 strategy board games
& Fixed states
& Perception, rule integration and local optimization
& Fixed-state question answering\\
\midrule
\textbf{DECE}
& \textbf{10 score-based Eurogames}
& 2--4 players\textsuperscript{*}; competition through game actions
& Long-term investment, persistent placements and multiple scoring routes
& Cross-game rank agreement; matched harness contrasts; resource frontiers\\
\bottomrule
\end{tabularx}
\par\vspace{3pt}
\begin{minipage}{\linewidth}\footnotesize
\textsuperscript{*}DECE evaluates three-player matches in all ten games, with supplementary
two-/four-player matches in Splendor and Brass, and four-player matches in
7 Wonders.
\end{minipage}
\end{table}


\section{DECE: Benchmark Design and Evaluation Protocol} 
\label{sec:dece}
\label{sec:suite}
\subsection{Game suite and strategic diversity} 
\label{sec:background}
DECE selects ten score-based Eurogames that share persistent strategic
commitments and competition over limited opportunities, while differing in how
current decisions shape future options. Table~\ref{tab:games} groups them
by three dominant, overlapping demands: building future capacity, making
persistent spatial commitments, and coordinating multiple action, resource,
and scoring systems. These differences allow the suite to test whether
competitive advantages recur across related strategic settings or remain
specific to individual games. The games also vary in interaction and
information structure: players compete entirely through legal game actions,
without dialogue or negotiation, and decisions may involve public or private
information as well as sequential or simultaneous choices.

\begin{figure}[!htbp]
\centering
\includegraphics[width=\linewidth]{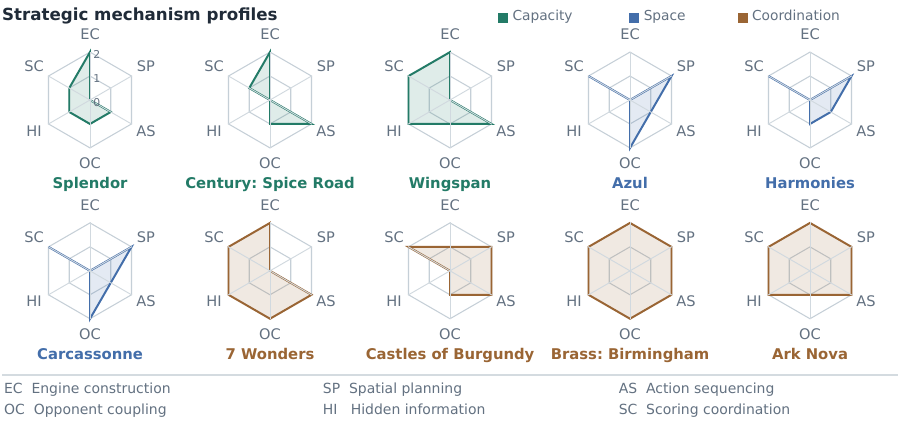}
\caption{\textbf{Complementary strategic demands across ten Eurogames.}
Each game is annotated along six dimensions. Values 0/1/2 denote absent/limited/prominent
mechanisms; colors indicate overlapping strategic emphases. 
Appendix~\ref{app:game-profiles} gives the rubric and evidence.}
\label{fig:game-profiles}
\end{figure}

Figure~\ref{fig:game-profiles} characterizes this diversity along six
rule-based dimensions: engine construction, spatial planning, action
sequencing, opponent coupling, hidden information, and scoring coordination.
Values 0/1/2 denote absent, limited, or prominent mechanisms.
These annotations organize descriptive cross-game comparisons
(Appendix~\ref{app:game-profiles}). The suite also spans substantially different interaction
lengths, with median whole-match decision counts ranging from 54 in
\emph{7 Wonders} to 473 in \emph{Ark Nova}. These counts summarize interaction
length at each interface's action granularity (Appendix~\ref{app:horizon-reporting}).

\begin{table}[!htbp]
\centering\small
\setlength{\tabcolsep}{4pt}
\caption{The selected ten-game roster and complementary strategic demands.
Groups and colors follow Figure~\ref{fig:game-profiles}.
Player ranges describe the selected competitive multiplayer scope;
solo modes are excluded.
Stars denote introductory ($\star$), intermediate ($\star\star$), and
advanced ($\star\star\star$) design tiers.}
\label{tab:games}
\begin{tabularx}{\linewidth}{@{}llccX@{}}
\toprule
Group & Game & Tier & Players & Distinctive strategic demand\\
\midrule
\textcolor[HTML]{237B67}{\textbf{Capacity}} & Splendor & $\star$ & 2--4 & Shared-market timing; engine investment\\
& Century: Spice Road & $\star$ & 2--5 & Resource conversion; hand cycling\\
& Wingspan & $\star\star$ & 2--5 & Tableau synergies; activation order\\
\midrule
\textcolor[HTML]{436EAA}{\textbf{Spatial}} & Azul & $\star$ & 2--4 & Drafting; spatial scoring; denial\\
& Harmonies & $\star\star$ & 2--4 & Three-dimensional pattern construction\\
& Carcassonne & $\star$ & 2--5 & Shared topology; follower commitment\\
\midrule
\textcolor[HTML]{9A6534}{\textbf{Coordination}} & 7 Wonders & $\star\star$ & 3--7 & Hidden drafting; simultaneous choices\\
& The Castles of Burgundy & $\star\star\star$ & 2--4 & Dice allocation; timed area completion\\
& Brass: Birmingham & $\star\star\star$ & 2--4 & Shared networks; finance; two-era economy\\
& Ark Nova & $\star\star\star$ & 2--4 & Action strength; coupled development tracks\\
\bottomrule
\end{tabularx}
\end{table}

\subsection{Unified game interface and participant abstraction} 

All games share an interface for setup, seat-specific observations, legal
actions, state transitions, and terminal rankings. The engine enforces rules
and scoring. Simultaneous choices use a frozen state and advance only after
all required seats submit actions. Public information, private state, and
unrevealed randomness remain separate. MCP and native adapters expose this
interface; persisted states and execution records support auditing, replay,
and resume
(Appendix~\ref{app:integration-flow}).

Each participant specifies a model snapshot, execution harness, prompts,
tools, memory scope, inference settings, and resource budget. The shared
interface permits different internal histories and orchestration strategies.
The primary panel contains 12 configurations spanning nine model identities
and four harnesses (Appendix~\ref{app:identities}). Comparisons evaluate these
complete configurations.

\subsection{Competitive outcomes and resource accounting}

We convert terminal placements into pairwise wins, assigning half credit
for ties, and fit regularized Bradley--Terry ratings on an Elo scale
\citep{bradley1952rank}:
\begin{equation}
p_{ij}=\frac{1}{1+10^{(E_j-E_i)/400}},
\qquad E_i=1500+\frac{400}{\ln 10}\theta_i,
\quad \sum_i\theta_i=0.
\label{eq:elo-scale}
\end{equation}

Ratings express relative strength and are centered within each comparison
population. Appendix~\ref{app:elo-cost}
specifies fitting and uncertainty.

Pairwise placement score (PWS) is the fraction of opponents outranked, with
half credit for ties. Mean PWS weights the ten games equally. Additional
player counts are analyzed separately, and uncertainty uses whole-match
resampling (Appendix~\ref{app:pws-diagnostic}).

We pair three-player outcomes with input-plus-output tokens and
API-equivalent cost under a fixed pricing schedule. Per-game Elo--cost and
Elo--token frontiers retain configurations for which no alternative achieves
at least as high performance with no greater resource use
(Appendix~\ref{app:elo-cost}).

\section{Experimental Evaluation}
\label{sec:experiments}
\subsection{Cohorts and uncertainty}
The main comparison contains 480 three-player matches across ten games:
144 from the original Splendor, 7 Wonders, and Brass cohort, plus 336 from
the seven-game expansion. Each game contributes 48 matches and 12
appearances per participant, balanced across seats. Another 252 two- or
four-player matches support player-count analysis, giving 732 native-system
matches in total. All outcomes reconcile with schedules and final records
(Appendix~\ref{app:reproducibility}). Table~\ref{tab:performance-compact}
reports the three-player cohort; PWS means weight games equally.
Its column groups follow Figure~\ref{fig:game-profiles}.
Pointwise 95\% intervals use 5,000 whole-match bootstrap draws per game,
recomputing PWS and refitting Elo while retaining the ten games
(Appendix~\ref{app:completed-inference}).

We completed 180 matches for three non-LLM algorithms---Random, Heuristic,
and ISMCTS---in separate 60-match cohorts. Their PWS scores gauge their approximate
competitive standing relative to the 12 LLM participants
(Table~\ref{tab:performance-compact}). Each cohort contains six three-player
matches per game against two LLM opponents. The three cohorts use
identical games, rules, setup seeds, opponent configurations, and seat
assignments; only the baseline algorithm changes
(Appendix~\ref{app:nonllm-baselines}).


Two independent studies cross OpenCode and Qwen Code in Splendor and
7 Wonders. The 72-match bridge pairs DSV4-Flash with Qwen3.8-Flash across
nine scenarios per game; the 48-match within-Qwen-family study pairs Qwen3.8-Flash with
Qwen3.8-Max across six. Within each study, those models under OpenCode serve
as anchors. Each scenario fixes the setup seed and focal seat across four
matches, with balanced seats. Pointwise intervals resample whole scenarios.
Each study retains its own anchors and analysis, separate from the native
ratings (Appendices~\ref{app:bridge} and~\ref{app:qwen-harness-pilot}).

\subsection{RQ1: Cross-game performance profiles}
\begin{table}[!htbp]
\centering\footnotesize
\caption{Three-player PWS (\%) across ten games.
IDs 1--12 use the 480-match main cohort; Mean weights games equally.
Bold marks the highest main-cohort estimate per game. IDs 13--15 are separate-cohort non-LLM baselines (Appendix~\ref{app:nonllm-baselines}).}
\label{tab:performance-compact}
\setlength{\tabcolsep}{1.4pt}
\begin{tabular*}{\linewidth}{@{\extracolsep{\fill}}lr*{11}{r}@{}}
\toprule
& & \multicolumn{3}{c}{Capacity} & \multicolumn{3}{c}{Space} & \multicolumn{4}{c}{Coordination} & \\
\cmidrule(lr){3-5}\cmidrule(lr){6-8}\cmidrule(lr){9-12}
Participant & ID & SP & CE & WI & AZ & HA & CA & 7W & BU & BR & AN & Mean \\
\midrule
GPT-5.6-Sol / Codex & 1 & 66.7 & 66.7 & \textbf{83.3} & 66.7 & 75.0 & 66.7 & 66.7 & 70.8 & \textbf{95.8} & \textbf{95.8} & 75.4 \\
GPT-5.6-Sol / OpenCode & 2 & 54.2 & \textbf{87.5} & 75.0 & 66.7 & 79.2 & 81.2 & \textbf{75.0} & \textbf{87.5} & 66.7 & 83.3 & \textbf{75.6} \\
GPT-5.6-Terra / Codex & 3 & 37.5 & 70.8 & 45.8 & 33.3 & 37.5 & 39.6 & 41.7 & 41.7 & 79.2 & 75.0 & 50.2 \\
GPT-5.6-Luna / Codex & 4 & 20.8 & 12.5 & 16.7 & 37.5 & 37.5 & 29.2 & 25.0 & 41.7 & 33.3 & 41.7 & 29.6 \\
DSV4-Pro / DSH & 5 & 58.3 & 58.3 & 62.5 & 58.3 & 58.3 & 66.7 & 66.7 & 58.3 & 58.3 & 58.3 & 60.4 \\
DSV4-Pro / OpenCode & 6 & 58.3 & 50.0 & 37.5 & 45.8 & 62.5 & 50.0 & 62.5 & 62.5 & 54.2 & 58.3 & 54.2 \\
DSV4-Flash / DSH & 7 & 50.0 & 41.7 & 50.0 & 54.2 & 29.2 & 33.3 & 50.0 & 50.0 & 37.5 & 33.3 & 42.9 \\
Qwen3.8-Max / OpenCode & 8 & 50.0 & 62.5 & 70.8 & 45.8 & 79.2 & 58.3 & 54.2 & 37.5 & 54.2 & 70.8 & 58.3 \\
Qwen3.8-Max / Qwen Code & 9 & \textbf{70.8} & 75.0 & \textbf{83.3} & \textbf{79.2} & \textbf{89.6} & \textbf{91.7} & 41.7 & 79.2 & 54.2 & 37.5 & 70.2 \\
Qwen3.7-Plus / OpenCode & 10 & 20.8 & 0.0 & 20.8 & 16.7 & 4.2 & 18.8 & 41.7 & 4.2 & 29.2 & 4.2 & 16.0 \\
GLM-5.2 / OpenCode & 11 & 62.5 & 50.0 & 29.2 & 45.8 & 41.7 & 35.4 & 54.2 & 41.7 & 29.2 & 29.2 & 41.9 \\
Kimi-K2.7-Code / OpenCode & 12 & 50.0 & 25.0 & 25.0 & 50.0 & 6.2 & 29.2 & 20.8 & 25.0 & 8.3 & 12.5 & 25.2 \\
\midrule
Random (separate cohort) & 13 & 0.0 & 0.0 & 0.0 & 0.0 & 0.0 & 0.0 & 0.0 & 0.0 & 8.3 & 0.0 & 0.8 \\
Heuristic (separate cohort) & 14 & 50.0 & 0.0 & 0.0 & 8.3 & 0.0 & 0.0 & 8.3 & 8.3 & 16.7 & 8.3 & 10.0 \\
ISMCTS (separate cohort) & 15 & 66.7 & 66.7 & 0.0 & 75.0 & 70.8 & 16.7 & 16.7 & 41.7 & 83.3 & 8.3 & 44.6 \\
\bottomrule
\end{tabular*}

\par\vspace{3pt}
\begin{minipage}{\linewidth}\scriptsize
SP: Splendor; CE: Century; WI: Wingspan; AZ: Azul; HA: Harmonies; CA: Carcassonne;\\
7W: 7 Wonders; BU: Burgundy; BR: Brass; AN: Ark Nova.
\end{minipage}
\end{table}

\paragraph{Aggregate rankings generalize across games.}
Nine-game mean PWS rankings agree with the held-out game at mean Kendall
$\tau_b=0.664$ and 83.9\% pair-order accuracy; per-game $\tau_b$ ranges from
0.489 to 0.862.
These results support recurring cross-game rankings alongside local
variation; Appendix~\ref{app:completed-inference} gives ranking uncertainty
and split-half diagnostics.

\paragraph{Broad strength and concentrated advantages.}
GPT-5.6-Sol/OpenCode and GPT-5.6-Sol/Codex lead equal-game PWS at 75.6\% and
75.4\%; both remain above 50\% in every game and within the top five by
three-player Elo. Their 0.2 pp difference has a whole-match interval of
[$-$8.6, 9.3]. Qwen3.8-Max/Qwen Code has lower
mean PWS (70.2\%) but leads or ties five game columns. DSV4-Pro/DSH has no
column lead, yet stays between 58.3\% and 66.7\% across games. These profiles
distinguish broad competitiveness from local peaks and weaknesses.

\paragraph{Advantages recur across selected strategic settings.}
Qwen3.8-Max/Qwen Code ranks first by Elo in Azul, Harmonies, and Carcassonne,
and second in Burgundy. All four combine prominent spatial constraints
with public current holdings under the rubric in
Figure~\ref{fig:game-profiles}; their equal-game PWS is 84.9\%.
GPT-5.6-Terra/Codex shows a different concentration: it ranks third, second,
and fourth in Century, Brass, and Ark Nova (75.0\% mean PWS), but eleventh,
tenth, and seventh in Azul, Harmonies, and Carcassonne (36.8\%). Its ten-game mean of 50.2\%
conceals this separation. Together with the Qwen3.8-Max profile, this illustrates how similar aggregate summaries can arise from markedly different game-level performance patterns.

\paragraph{A single mechanism does not summarize the profile.}
Brass and Ark Nova also have prominent spatial constraints, but Qwen3.8-Max/Qwen Code ranks sixth and eighth there. Conversely, GPT-5.6-Terra/Codex's
strong results in several economic systems do not extend to all
engine-building games: it ranks tenth in Splendor and sixth in Wingspan.
Averaging GPT-5.6-Terra over all spatial games gives 51.0\% PWS, versus 49.0\%
elsewhere, masking substantial differences within that group. These
counterexamples motivate examining combinations of commitments, information,
and action dependencies rather than assigning a model one domain label.

\paragraph{Configuration advantages across game groups.}
The preceding examples suggest that some configuration differences may recur across games sharing particular rule properties. We therefore ask whether same-model harness differences are concentrated within a rule-defined subset of games rather than cancelling when averaged across the full suite.
For each model, $\delta_g$ is its PWS under harness A minus its PWS under
OpenCode in game $g$, measured in percentage points.
Group $G$ comprises Azul, Harmonies, Carcassonne, and Burgundy;
$H$ contains the other six games.
All four games have spatial planning = 2 and hidden information = 0
under the rule annotations in Figure~\ref{fig:game-profiles}.
This criterion differs from the dominant strategic emphases used in
Tables~\ref{tab:games} and~\ref{tab:performance-compact}, where Burgundy
belongs to Coordination.

Let $\overline\Delta_G$ and $\overline\Delta_H$ denote the equal-game means
of $\delta_g$ in each group. Their difference,
$\Gamma=\overline\Delta_G-\overline\Delta_H$, measures how the average
configuration advantage differs between groups.
A positive $\Gamma$ indicates that harness A has a larger average advantage over OpenCode in the public-spatial group than in the remaining games.
These contrasts describe configuration performance across the rule-defined
game groups under the native opponent samples.
Appendix~\ref{app:specialization} gives the interval and game-deletion procedures.

\begin{table}[!htbp]
\centering\footnotesize
\setlength{\tabcolsep}{3pt}
\caption{Same-model configuration differences across ten games (PWS, percentage points).
(a) Harness A minus OpenCode for each game.
(b) Equal-game group means, their difference $\Gamma$ with pointwise 95\% intervals,
and the range of $\Gamma$ after omitting one game at a time.
Comparisons use the 480-match native cohort.}
\label{tab:configuration-concentration}
\definecolor{deceGain}{HTML}{267C8D}
\definecolor{deceLoss}{HTML}{B55F3F}
\begingroup
\setlength{\tabcolsep}{2pt}
\renewcommand{\arraystretch}{1.10}
\noindent\textbf{(a) Game-wise differences: harness A minus OpenCode}\par\smallskip
\begin{tabular*}{\linewidth}{@{\extracolsep{\fill}}ll*{4}{r}@{\hspace{9pt}}*{6}{r}@{}}
\toprule
 & & \multicolumn{4}{c}{Public spatial ($G$)} & \multicolumn{6}{c}{Other games ($H$)} \\
\cmidrule(lr){3-6}\cmidrule(l){7-12}
Model & Harness A & AZ & HA & CA & BU & SP & CE & WI & 7W & BR & AN \\
\midrule
Qwen3.8-Max & Qwen Code & \cellcolor{deceGain!23}$+33.3$ & \cellcolor{deceGain!7}$+10.4$ & \cellcolor{deceGain!23}$+33.3$ & \cellcolor{deceGain!29}$+41.7$ & \cellcolor{deceGain!15}$+20.8$ & \cellcolor{deceGain!9}$+12.5$ & \cellcolor{deceGain!9}$+12.5$ & \cellcolor{deceLoss!9}$-12.5$ & $0.0$ & \cellcolor{deceLoss!23}$-33.3$ \\
GPT-5.6-Sol & Codex & $0.0$ & \cellcolor{deceLoss!3}$-4.2$ & \cellcolor{deceLoss!10}$-14.6$ & \cellcolor{deceLoss!12}$-16.7$ & \cellcolor{deceGain!9}$+12.5$ & \cellcolor{deceLoss!15}$-20.8$ & \cellcolor{deceGain!6}$+8.3$ & \cellcolor{deceLoss!6}$-8.3$ & \cellcolor{deceGain!20}$+29.2$ & \cellcolor{deceGain!9}$+12.5$ \\
DSV4-Pro & DSH & \cellcolor{deceGain!9}$+12.5$ & \cellcolor{deceLoss!3}$-4.2$ & \cellcolor{deceGain!12}$+16.7$ & \cellcolor{deceLoss!3}$-4.2$ & $0.0$ & \cellcolor{deceGain!6}$+8.3$ & \cellcolor{deceGain!18}$+25.0$ & \cellcolor{deceGain!3}$+4.2$ & \cellcolor{deceGain!3}$+4.2$ & $0.0$ \\
\bottomrule
\end{tabular*}
\par\medskip
\noindent\textbf{(b) Group concentration and sensitivity to game composition}\par\smallskip
\begin{tabular*}{\linewidth}{@{\extracolsep{\fill}}lrrrr@{}}
\toprule
Model & \shortstack{Public spatial\\$\overline\Delta_G$ (4 games)} & \shortstack{Other games\\$\overline\Delta_H$ (6 games)} & \shortstack{Group difference\\$\Gamma$ [95\% interval]} & \shortstack{Delete-one-game\\range of $\Gamma$} \\
\midrule
Qwen3.8-Max & $+29.7$ & $0.0$ & $\mathbf{+29.7}\;[11.9, 46.9]$ & $[+23.0, +36.1]$ \\
GPT-5.6-Sol & $-8.9$ & $+5.6$ & $\mathbf{-14.4}\;[-32.6, 4.2]$ & $[-19.7, -9.7]$ \\
DSV4-Pro & $+5.2$ & $+6.9$ & $\mathbf{-1.7}\;[-21.8, 18.1]$ & $[-5.6, +1.9]$ \\
\bottomrule
\end{tabular*}
\endgroup

\par\vspace{3pt}
\begin{minipage}{\linewidth}\scriptsize
Game abbreviations follow Table~\ref{tab:performance-compact}.
Teal/orange denotes positive/negative differences on a common $\pm50$ pp scale.
Deletion ranges describe sensitivity to game composition, not confidence intervals.
\end{minipage}
\end{table}
Table~\ref{tab:configuration-concentration} separates recurring advantages
from cancellation across games. Qwen3.8-Max/Qwen Code exceeds OpenCode in
all four public-spatial games (+10.4 to +41.7 pp). Its zero mean advantage
elsewhere masks gains in Splendor (+20.8) and losses in Ark Nova ($-33.3$).
The between-group difference is +29.7 pp [11.9, 46.9] and ranges from
+23.0 to +36.1 when one game is omitted.

The other pairs have different profiles: GPT-5.6-Sol/Codex has three negative
and one zero contrast in the public-spatial group, while DSV4-Pro/DSH has
two positive and two negative contrasts. Game deletion preserves the sign
of the Qwen3.8-Max and GPT-5.6-Sol group differences, whereas DSV4-Pro's range
crosses zero. These composition checks complement the sampling intervals
conditional on the selected groups.
Appendix~\ref{app:specialization} gives further sensitivity checks;
RQ2 examines separate model panels under matched scenarios.


These comparisons fix three players. Appendix~\ref{app:player-count-panel}
reports supplementary player-count results and their opponent-composition scope.

\subsection{RQ2: Performance across harnesses}
The native-cohort comparisons in Section 4.2 suggest that the same registered model can exhibit different game-level profiles across harnesses, but these comparisons involve different opponent samples. We therefore conduct two matched studies in Splendor and 7 Wonders, fixing the setup seed, focal seat, and anchor configurations within each scenario.

\begin{figure}[!t]
\centering
\includegraphics[width=\linewidth]{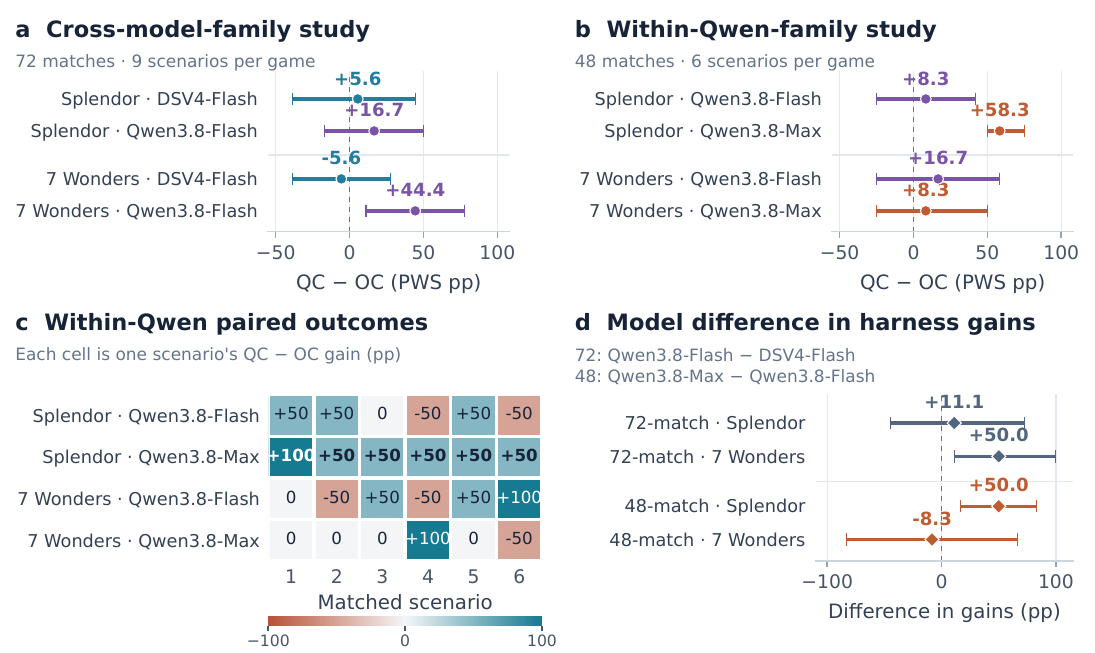}
\caption{Harness contrasts in two independent studies.
(a,b) Paired PWS gains, Qwen Code (QC) minus OpenCode (OC).
(c) All six scenario gains per model--game cell in the 48-match within-Qwen-family study;
columns match within each game.
(d) Differences in gains: Qwen3.8-Flash minus DSV4-Flash (0731 snapshot; 72 matches),
and Qwen3.8-Max minus Qwen3.8-Flash (48).
Whiskers: pointwise 95\% scenario-bootstrap intervals,
without multiplicity correction.}
\label{fig:matched-bridge}
\end{figure}

\paragraph{Cross-model-family contrasts.}
Against the shared anchors in the 72-match bridge, DSV4-Flash leads Qwen3.8-Flash under OpenCode
(58.3\% versus 33.3\% pooled PWS); under Qwen Code, the point estimates
reverse (58.3\% versus 63.9\%). Their difference in harness gains is
+30.6 pp [$-$5.6, 69.4].
Qwen3.8-Flash's largest gain is in 7 Wonders (+44.4 pp;
Figure~\ref{fig:matched-bridge}a).

\paragraph{Within-Qwen-family contrasts.}
In the 48-match within-Qwen-family study, Qwen3.8-Max's Splendor PWS rises from 33.3\% under OpenCode
to 91.7\% under Qwen Code, a +58.3 pp contrast [50.0, 75.0].
All six matched scenarios favor Qwen Code (Figure~\ref{fig:matched-bridge}c), whereas Qwen3.8-Flash improves by only +8.3 pp in the same game. The corresponding difference in harness gains is +50.0 pp in Splendor but \ensuremath{-}8.3 pp in 7 Wonders (Figure~\ref{fig:matched-bridge}d), indicating that harness effects can vary across both models and games. The six positive Qwen3.8-Max/Splendor pairs yield a two-sided sign-test $p=0.03125$, or $0.125$ after correction for four model–game contrasts (Appendix~\ref{app:qwen-harness-pilot}). We therefore view this as a strong local contrast that motivates replication, rather than evidence for a general model--harness interaction.

\FloatBarrier
\subsection{RQ3: Performance--resource trade-offs}
\begin{figure}[!t]
\centering
\includegraphics[width=\linewidth]{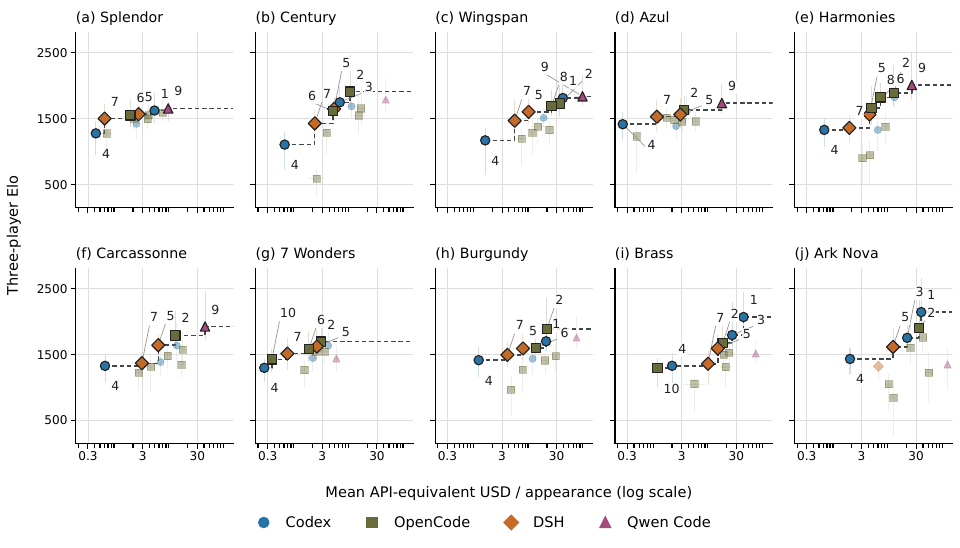}

\caption{Elo--cost Pareto frontiers by game (480 three-player matches;
IDs from Table~\ref{tab:performance-compact}).
Outlined points are nondominated; staircases show maximum estimated Elo
per mean-cost budget. Axes are shared; whiskers: pointwise 95\%
whole-match intervals. Stability and resources: Appendix~\ref{app:elo-cost};
token frontiers: Figure~\ref{fig:elo-token}.}

\label{fig:elo-cost}
\end{figure}

For each game, we compare three-player Elo with mean API-equivalent cost
per appearance under fixed prices (Figure~\ref{fig:elo-cost}).
All 480 matches contribute; 5,000 whole-match resamples recompute both
axes and frontier membership.

\paragraph{Efficient configurations depend on the game.}
Point-estimate frontiers contain five to seven configurations.
GPT-5.6-Luna/Codex and DSV4-Pro/DSH appear in all ten games, while
GPT-5.6-Sol/OpenCode appears in nine.
Qwen3.8-Max/Qwen Code lies on five frontiers, reaching the high-cost end
in Wingspan (Elo 1837, \$89.25),
but is dominated in Burgundy and Ark Nova. GPT-5.6-Sol/Codex also lies on five
frontiers and reaches the highest Elo in Brass (2068, \$40.53 per appearance)
and Ark Nova (2143, \$37.20).
For a given mean-cost budget, the staircase identifies the highest estimated
Elo among affordable configurations. Moving along it shows where a higher
budget changes the preferred configuration. Coverage across games describes
how often a configuration is nondominated; its location within each game
shows the cost and playing strength it offers.

\paragraph{Frontier membership has unequal sampling stability.}
GPT-5.6-Sol/Codex remains nondominated in 90.3\% of Brass resamples and 94.0\% of
Ark Nova resamples, compared with 44.3\% in Splendor. Qwen3.8-Max/Qwen Code
remains nondominated in 46.8\% of Wingspan resamples, 69.8\% in Harmonies,
and 75.0\% in Carcassonne.
These frequencies indicate how often a configuration remains a candidate
when the observed matches are resampled. They distinguish frequently
retained options from positions that change with the sampled outcomes.

\paragraph{Token frontiers separate usage from pricing.}
Using mean input-plus-output tokens yields one to six frontier
configurations per game (Figure~\ref{fig:elo-token}).
GPT-5.6-Luna/Codex appears on three token frontiers, compared with ten
cost frontiers; DSV4-Pro/DSH appears on two, compared with ten.
Kimi-K2.7-Code/OpenCode shows the reverse pattern, appearing on six token
frontiers and no dollar frontiers. The different candidate sets reflect
both token consumption and the fixed rates applied to input, cached input,
and output. Reporting both measures makes these sources of resource
efficiency visible.
Token usage is measured in each model's native tokenizer
(Appendix~\ref{app:elo-cost}).

\section{Discussion and Conclusion}
\label{sec:discussion}
DECE finds both cross-game ranking agreement and local differences in
competitive performance. The games share persistent investments, contested
opportunities, and alternative scoring routes, while differing in rules
and information structure. The shared roster reveals where
advantages recur and where they reverse. Aggregate rankings summarize
broad competitiveness, while game-level profiles identify settings in
which a configuration becomes more or less effective.

\paragraph{Complete systems as the evaluation unit.}
Matched comparisons connect these profiles to agent execution. The same
model can perform differently across harnesses under matched scenarios.
Competitive results characterize the complete model--harness configuration.
Reporting tools,
context management, and execution settings makes that system explicit.
Cross-game comparisons show whether configuration gains recur or
concentrate in particular strategic settings.

\paragraph{From profiles to configuration selection.}

The resource frontiers identify candidate configurations for a game and budget.
Future work can select configurations from game rules and results on
training games. Evaluation should compare against the best generalist
selected on those games, keeping target-game outcomes out of tuning.
Applying the same budget to both methods would test whether game-specific
selection improves competitive performance at a given resource cost.
Token and dollar budgets would test distinct deployment constraints.

\paragraph{Limitations.}
The native and matched studies evaluate complete configurations within their
respective game and opponent panels. Component interventions, combined with
action--outcome analyses, can test how planning and memory contribute to
these performance differences.

DECE combines executable Eurogames, common tournament protocols, and
resource measurements linked to competitive outcomes. Together, these
elements support evaluation that reports where complete agent configurations
perform well, how execution choices change their results, and what
resources they require.

\clearpage
\bibliography{references}
\bibliographystyle{references}
\appendix
\clearpage
\section{Implementation Scope and Game Documentation}
\label{app:protocol-details}
\subsection{Implemented rules and information scope}
\label{app:game-catalog}
All ten games contribute to the common three-player cohort.
Table~\ref{tab:implementation-scope} lists the implemented rule versions and
configuration details. Descriptions of investments, scoring, and interaction
appear with their rule evidence in Appendix~\ref{app:game-profiles}.

\begin{table}[!htbp]
\centering\scriptsize
\setlength{\tabcolsep}{3pt}
\renewcommand{\arraystretch}{1.12}
\caption{Implemented game configurations. Rule versions come from engine
metadata and experiment records; expansion options were checked against
all 336 final databases.}
\label{tab:implementation-scope}
\begin{tabularx}{\linewidth}{@{}>{\raggedright\arraybackslash}p{.19\linewidth}>{\raggedright\arraybackslash}p{.43\linewidth}>{\raggedright\arraybackslash}X@{}}
\toprule
Game & Rules edition & Configuration scope\\
\midrule
Splendor & Classic base rules & Developments, reservations, and nobles.\\
Century: Spice Road & Plan B base rules (2017) & Merchant hands public in DECE.\\
Wingspan & Stonemaier base English rules & Base English set; green goal board.\\
Azul & Next Move base rules (2018) & Colored-wall variant.\\
Harmonies & Libellud base rules (2024) & Board A; nature spirits disabled.\\
Carcassonne & Base rules with farmers & Base game with farmers.\\
7 Wonders & Second Edition (2020) & Base game; three ages and sealed drafting.\\
The Castles of Burgundy & Ravensburger base rules (2019) & Base game.\\
Brass: Birmingham & Roxley rules (November 2018) & Canal and rail eras.\\
Ark Nova & Base rules v1.2; FAQ v2 & Map 0; base-game content.\\
\bottomrule
\end{tabularx}
\end{table}


\paragraph{Information and action representation.}
The POSG framework accommodates private observations, public states,
and sequential or simultaneous decisions \citep{hansen2004posg}.
DECE exposes a seat-specific observation and legal action identifiers.
Public holdings, private information, unrevealed draws, and sealed choices
are recorded as distinct properties. The interface has no inter-player
dialogue channel. The game engine applies rules and determines outcomes;
persisted states and execution records support replay and resume.
Terminal outcomes undergo the checks in Appendix~\ref{app:reproducibility}.

\paragraph{Rules sources and attribution.}
Publisher documentation supplies the game-design references
\citep{rules_splendor,rules_century,rules_azul,rules_carcassonne,
rules_seven_wonders,rules_harmonies,rules_wingspan,rules_burgundy,
rules_brass,rules_ark_nova}; recorded configurations determine the evaluated
scope.
Cover thumbnails in Figure~\ref{fig:arena-overview} identify the games;
artwork remains with the respective rights holders.

\subsection{Participant identities and recorded settings}
\label{app:identities}
GPT-5.6-Sol, GPT-5.6-Terra, and GPT-5.6-Luna identify the three GPT-5.6 variants in
the evaluated participant panel \citep{openai2026gpt56systemcard}.
The Qwen configurations use Qwen3.8-Max \citep{qwen2026qwen38max} and
Qwen3.7-Plus \citep{alibaba2026qwen37plus}.
DSV4-Pro and DSV4-Flash denote the registered DeepSeek V4 Pro 0813 and V4 Flash
0731 snapshots of the DeepSeek-V4 family \citep{deepseek2026v4}.
GLM-5.2's official model card \citep{zai2026glm52} recommends
the GLM-5 technical report \citep{glm2026glm5}.
The Kimi configuration uses Kimi-K2.7-Code \citep{moonshot2026kimik27code}.
The separate harness bridge uses Qwen3.8-Flash
\citep{alibaba2026qwen38flash} and the DSV4-Flash 0731 snapshot.
These reports, model cards, and release documents describe the corresponding
models or families; the recorded configurations identify the evaluated
snapshots. Each model--harness configuration remains one participant.

\begin{table}[!htbp]
\centering\footnotesize
\caption{Registered model strings and effort fields in the completed
seven-game expansion. IDs match the primary participant panel in
Table~\ref{tab:performance-compact}.}
\label{tab:participant-settings}
\begin{tabular*}{\linewidth}{@{\extracolsep{\fill}}rlll@{}}
\toprule ID & Recorded model string & Harness & Effort \\
\midrule
1 & \texttt{gpt-5.6-sol} & Codex & xhigh \\
2 & \texttt{openai/gpt-5.6-sol} & OpenCode & xhigh \\
3 & \texttt{gpt-5.6-terra} & Codex & xhigh \\
4 & \texttt{gpt-5.6-luna} & Codex & xhigh \\
5 & \texttt{deepseek-v4-pro-0813} & DSH & max \\
6 & \texttt{deepseek-v4-pro-0813} & OpenCode & max \\
7 & \texttt{deepseek/deepseek-v4-flash-0731} & DSH & max \\
8 & \texttt{qwen3.8-max} & OpenCode & max \\
9 & \texttt{qwen3.8-max} & Qwen Code & max \\
10 & \texttt{qwen3.7-plus} & OpenCode & max \\
11 & \texttt{glm-5.2} & OpenCode & high \\
12 & \texttt{kimi-k2.7-code} & OpenCode & none \\
\bottomrule
\end{tabular*}

\end{table}
\FloatBarrier

All 336 expansion results use the model strings and effort settings
shown in Table~\ref{tab:participant-settings}. Controller retries and
fallback actions are disabled. The matched studies use
their own configurations
(Appendix~\ref{app:bridge}).

\clearpage
\subsection{Integration and execution}
\label{app:integration-flow}
\begin{figure}[!htbp]
\centering
\includegraphics[width=\linewidth]{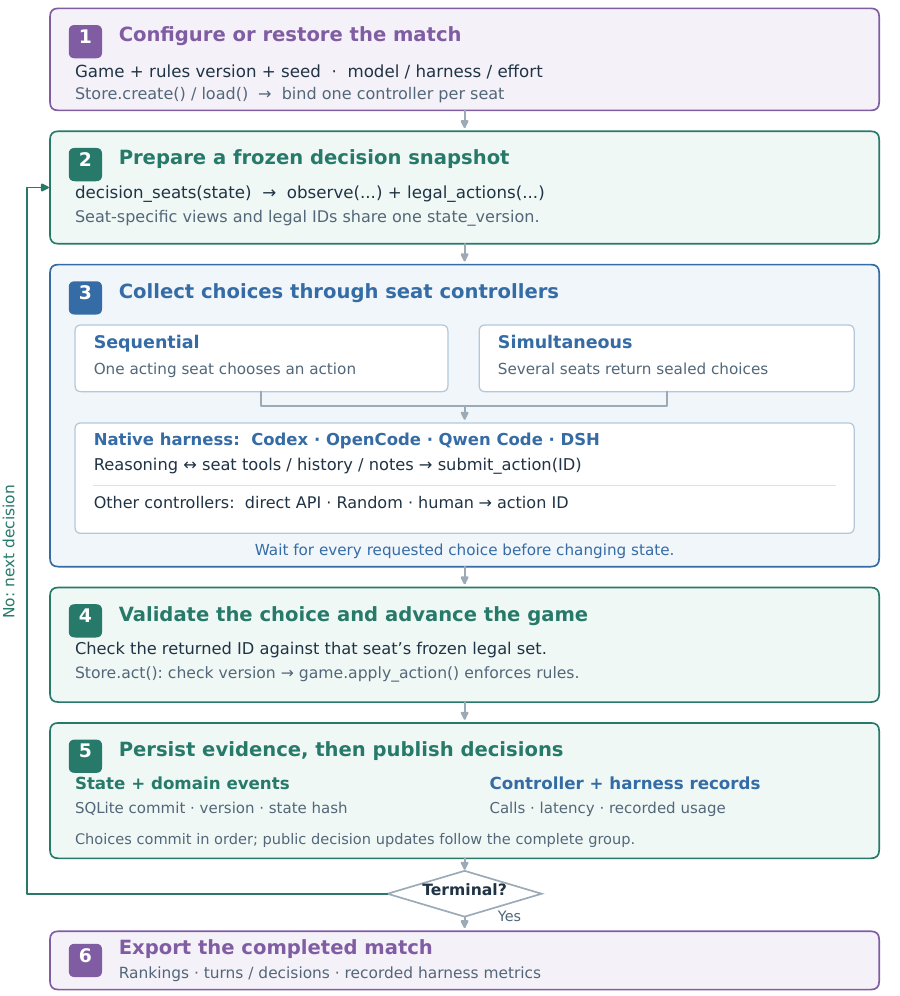}
\caption{\textbf{Integration and execution in DECE.} The numbered stages
follow one match; the loop repeats while play remains active. Controllers
receive seat-specific views and return action IDs. The versioned game plugin
applies rules, and the store commits state and events. Simultaneous choices
are collected before any action is applied; public decision updates follow the complete
group. Recorded calls and usage support resource analysis.}
\label{fig:integration-flow}
\end{figure}

\clearpage
\section{Mechanism Profiles and Annotation Rubric}

\label{app:game-profiles}

Figure~\ref{fig:game-profiles} describes the selected games using six rule-based annotations. The authors assigned ordinal values from publisher rules and the corresponding observation interfaces. RQ1 uses these annotations to organize descriptive comparisons, with group definitions in Appendix~\ref{app:specialization}.

\paragraph{Annotation rubric.} Each axis uses the following anchors. A value of zero means that the defined mechanism is absent, not that a game requires no reasoning. Hidden information concerns current private holdings; uncertainty about future random draws is excluded. Spatial planning concerns geometry and connectivity, not the mere presence of rows in a tableau.

\begin{description}

\item[EC: Engine construction.] 0: No persistent production or action modifier; 1: Localized recurring benefit or modifier; 2: Acquired assets repeatedly change action yields, costs, or available conversions.

\item[SP: Spatial planning.] 0: No geometric adjacency or connectivity constraint; 1: Geometry affects a subsidiary mechanism; 2: Adjacency, connectivity, or packing is central to placement or scoring.

\item[AS: Action sequencing.] 0: No material ordering dependency; 1: Resource or placement prerequisites create ordering constraints; 2: Explicit reuse, activation chains, action-strength cycles, or staged production dependencies.

\item[OC: Opponent coupling.] 0: Independent boards and supplies; only final score comparison; 1: Shared offers, resource pools, or races; 2: Trade or shared infrastructure, contested ownership, neighbor military comparisons, or opponent-shaped compulsory leftovers.

\item[HI: Hidden information.] 0: Current decision-relevant holdings public; future random draws excluded; 1: Limited or optional private holdings; 2: Private hands or goals are a recurrent central part of decisions.

\item[SC: Scoring coordination.] 0: One principal additive scoring conversion; 1: Main scoring route plus a secondary bonus or threshold; 2: Overlapping, nonlinear, comparative, or staged scoring objectives.

\end{description}

\paragraph{Reading the profiles.} The colors identify three overlapping strategic emphases: building future capacity, committing space and pieces, and coordinating several systems. Opponent coupling follows the main competitive mechanism in the evaluated configuration. Wingspan's habitat order is represented under action sequencing. Century's information annotation follows DECE's public merchant-hand observation. Ark Nova receives OC=1 for its shared displays and project races.

\begin{table}[ht]\centering\small

\caption{Explicit values behind Figure~\ref{fig:game-profiles}. These ordinal mechanism annotations are not added or averaged.}

\label{tab:game-profile-values}

\begin{tabular}{lrrrrrr}\toprule Game & EC & SP & AS & OC & HI & SC\\\midrule

Splendor & 2 & 0 & 1 & 1 & 1 & 1\\

Century: Spice Road & 2 & 0 & 2 & 1 & 0 & 1\\

Wingspan & 2 & 0 & 2 & 1 & 2 & 2\\

Azul & 0 & 2 & 1 & 2 & 0 & 2\\

Harmonies & 0 & 2 & 1 & 1 & 0 & 2\\

Carcassonne & 0 & 2 & 1 & 2 & 0 & 2\\

7 Wonders & 2 & 0 & 2 & 2 & 2 & 2\\

Castles of Burgundy & 1 & 2 & 2 & 1 & 0 & 2\\

Brass: Birmingham & 2 & 2 & 2 & 2 & 2 & 2\\

Ark Nova & 2 & 2 & 2 & 1 & 2 & 2\\

\bottomrule\end{tabular}\end{table}

\paragraph{Rule evidence.} The following notes explain every coordinate in axis order.

\begingroup\small

\paragraph{Splendor.} \textbf{EC}: Development bonuses permanently reduce later purchase costs. \textbf{SP}: Market rows and card tableaux impose no geometric scoring constraint. \textbf{AS}: Gem collection and discount prerequisites order purchases. \textbf{OC}: Shared cards, gems, and nobles create races. \textbf{HI}: Reserved developments can remain private. \textbf{SC}: Prestige purchases are supplemented by noble thresholds. \citep{rules_splendor}.

\paragraph{Century: Spice Road.} \textbf{EC}: Acquired merchants expand reusable resource-conversion options. \textbf{SP}: Card market order is a price rule, not geometric planning. \textbf{AS}: Played merchants require a rest action before reuse. \textbf{OC}: Merchant and point-card markets are shared. \textbf{HI}: DECE exposes merchant hands; unknown future draws do not count here. \textbf{SC}: Point cards dominate, with coins and leftover spices adding points. \citep{rules_century}.

\paragraph{Wingspan.} \textbf{EC}: Birds improve habitat actions and add recurring powers. \textbf{SP}: Habitat rows determine activation; geometric adjacency is not scored. \textbf{AS}: Habitat activations chain powers after prerequisite food and egg investments. \textbf{OC}: Food supply, bird offers, and round-goal races are shared. \textbf{HI}: Bird hands and bonus cards create recurring private information. \textbf{SC}: Bird points, eggs, cached food, tucked cards, and goals interact. \citep{rules_wingspan}.

\paragraph{Azul.} \textbf{EC}: Wall tiles score but do not produce resources or stronger actions. \textbf{SP}: Pattern-line capacity and wall adjacency govern placement and points. \textbf{AS}: Pattern lines must fill before tiles transfer to the wall. \textbf{OC}: Drafts reshape compulsory leftovers and possible floor penalties. \textbf{HI}: Boards and draft pools are public; bag randomness is excluded. \textbf{SC}: Adjacency scoring, row/column/color bonuses, and penalties interact. \citep{rules_azul}.

\paragraph{Harmonies.} \textbf{EC}: Landscape and animal pieces score without increasing action productivity. \textbf{SP}: Hex adjacency and stack heights define habitats and terrain scores. \textbf{AS}: Required landscape patterns must precede animal placements. \textbf{OC}: Token groups and animal offers are shared. \textbf{HI}: Personal landscapes and selected animal cards are public. \textbf{SC}: Animal habitats overlap with several terrain-specific scoring patterns. \citep{rules_harmonies}.

\paragraph{Carcassonne.} \textbf{EC}: Features score and recycle followers but do not create a production engine. \textbf{SP}: Shared terrain connectivity determines legal placement and features. \textbf{AS}: Completions release followers; fields retain commitments until the end. \textbf{OC}: Merging features can create majority contests for ownership. \textbf{HI}: Placed tiles and followers are public; future tile draws are excluded. \textbf{SC}: Completed features, unfinished features, and field-city relationships differ. \citep{rules_carcassonne}.

\paragraph{7 Wonders.} \textbf{EC}: Production and commerce developments change future building access. \textbf{SP}: Neighbor identity matters, but there is no geometric placement puzzle. \textbf{AS}: Production, free-building chains, and age progression order investments. \textbf{OC}: Neighbor trade, military comparisons, and passed hands couple players. \textbf{HI}: Private draft hands and sealed selections are central. \textbf{SC}: Science sets, military comparisons, civic points, and commerce interact. \citep{rules_seven_wonders}.

\paragraph{Castles of Burgundy.} \textbf{EC}: Mines and knowledge tiles add localized income or rule modifiers. \textbf{SP}: Estate adjacency, terrain, and connected regions constrain placements. \textbf{AS}: Acquisition, storage, placement, and immediate bonus actions form chains. \textbf{OC}: Shared depots and completion races couple otherwise personal estates. \textbf{HI}: Dice, tiles, and estates are public; future replenishment is excluded. \textbf{SC}: Region size, completion phase, animals, goods, and bonuses interact. \citep{rules_burgundy}.

\paragraph{Brass: Birmingham.} \textbf{EC}: Industry investments raise recurring income and expand economic capacity. \textbf{SP}: Network connectivity governs resource access and expansion. \textbf{AS}: Financing, building, resource access, selling, and era changes form dependencies. \textbf{OC}: Players consume shared resources and use transport infrastructure. \textbf{HI}: Private location and industry cards constrain available builds. \textbf{SC}: Industry and link scoring are coordinated across canal and rail eras. \citep{rules_brass}.

\paragraph{Ark Nova.} \textbf{EC}: Sponsors, action upgrades, and zoo development improve later actions or income. \textbf{SP}: Enclosure packing and map constraints govern zoo construction. \textbf{AS}: Five action-card strength positions cycle as cards are used. \textbf{OC}: Shared displays, projects, and race timing dominate the selected profile. \textbf{HI}: Private animal/sponsor hands and final-scoring cards affect plans. \textbf{SC}: Appeal, conservation, project thresholds, and final scoring interact. \citep{rules_ark_nova}.

\endgroup

\FloatBarrier
\section{Comparison Scope and Horizon Reporting}
\label{app:horizon-reporting}
\subsection{Comparison-table scope and sources}
Table~\ref{tab:rq-coverage} distinguishes game scope, evaluated
interaction, decision demands, and evaluation focus. Decision demands
summarize rules and tasks; they are not measured capability scores.
DECE's interaction entry includes the supplementary player counts in
Table~\ref{tab:expanded-exposure}. Its cross-game and resource analyses
use the common three-player cohort.

\paragraph{Sources for Table~\ref{tab:rq-coverage}.}
GameBench covers spatial strategy, social deduction, and deck building
\citep[Section 3.2; Appendices F--G]{costarelli2024gamebench}.
MindGames reports two- through six-player settings
\citep[Section 3.2]{wang2026mindgames}; FinBoardBench evaluates four-player
financial board games \citep[Sections 2--3]{hu2026finboardbench}.
PTCG-Bench uses five decks within one two-player card game and includes
harness and resource analyses \citep[Section 4; Appendices B, E--F]{hua2026ptcg}.
From Code to Play evaluates generated state-evaluation functions in
search agents across twelve TAG games
\citep[Section V-G; Table VIII]{eberhardinger2024codetoplay}.
LudoBench evaluates questions about fixed states in five tabletop games
\citep{peper2026ludobench}.
The table uses GameBench v2, MindGames v1, FinBoardBench v1,
PTCG-Bench v2, From Code to Play v2, and the ICLR 2026 LudoBench release.

\subsection{Observed decision lengths}
Table~\ref{tab:dece-horizon} uses the same 480 three-player matches as the
primary performance matrix, with 48 matches per game. It summarizes the
recorded whole-match decision counter $D$, engine turns, and $D/3$ using
linear-interpolation quantiles. Median $D$ ranges from 54 in 7 Wonders to
473 in Ark Nova; Splendor and Brass have medians of 81 and 393.5.
These are observed counters for the evaluated systems and schedules.
\begin{table}[ht]
\centering\small
\setlength{\tabcolsep}{4pt}
\renewcommand{\arraystretch}{1.12}
\caption{Recorded decision and turn counts for all ten native games in the 480-match
three-player cohort. $D$ is the whole-match decision counter; brackets give
its interquartile range. Engine turns and $D/3$ are medians over matches.
Counter granularity is game-specific; $D/3$ is a within-match average over
seats, not an individual-player distribution.}
\label{tab:dece-horizon}
\begin{tabular*}{\linewidth}{@{\extracolsep{\fill}}lrrrr@{}}
\toprule
Game & Matches & $D$: median [IQR] & Engine turns & $D/3$ \\
\midrule
Splendor & 48 & 81 [77.2, 84] & 81 & 27 \\
Century & 48 & 129 [119.2, 135] & 129 & 43 \\
Wingspan & 48 & 370 [357.8, 411.2] & 370 & 123.3 \\
Azul & 48 & 70 [67.8, 72.2] & 70 & 23.3 \\
Harmonies & 48 & 166.5 [158.8, 171.2] & 166.5 & 55.5 \\
Carcassonne & 48 & 142 [142, 142] & 142 & 47.3 \\
7 Wonders & 48 & 54 [54, 55] & 18 & 18 \\
Burgundy & 48 & 281 [277, 284.5] & 75 & 93.7 \\
Brass: Birmingham & 48 & 393.5 [346.5, 416] & 54 & 131.2 \\
Ark Nova & 48 & 473 [456.5, 502.5] & 103 & 157.7 \\
\bottomrule
\end{tabular*}

\end{table}

The counters retain their game-specific source semantics. Splendor's
decision counts use player turns and omit extra noble-selection
decisions. In 7 Wonders, 18 synchronous drafting rounds correspond to
54--55 recorded decisions across three players. Brass has 54 player turns
but several accepted sub-decisions within a turn. In other games the recorded
turn counter may itself advance at the sub-decision level; equality of
turn and decision counts does not imply comparable strategic units.
A larger count can therefore reflect interface granularity as well as
sustained interaction.

Dividing $D$ by three gives the within-match average over seats before taking
the median. Actual individual-player trajectory lengths require accepted-action
logs and seat attribution. Controller decisions and model tool calls are
separate execution measures (Appendix~\ref{app:telemetry}). These counters document the evaluated interaction scope;
they do not measure strategic depth on a common scale across games.

\FloatBarrier
\section{Complete Results and Statistical Protocol}
\subsection{Ten-game three-player cohort}
\label{app:expanded-suite}
The primary native comparison contains 480 three-player matches across ten
games and 1,440 participant appearances. Each of 12 configurations appears
12 times per game, four times at each seat. Table~\ref{tab:expanded-exposure}
reports this common exposure and the additional player-count coverage.
\begin{table}[!htbp]
\centering\footnotesize
\caption{Native match coverage across all ten games. Appearance and seat
columns refer to the three-player cohort. The final columns contain the
252 supplementary two-/four-player matches; dashes mean no observed stratum.
Each three-player game has the same 12-configuration roster.}
\label{tab:expanded-exposure}
\setlength{\tabcolsep}{2pt}
\begin{tabular*}{\linewidth}{@{\extracolsep{\fill}}lrrrrr@{}}
\toprule
Game & 3p matches & Appear./config. & Each seat & 2p matches & 4p matches \\
\midrule
Splendor & 48 & 12 & 4 & 72 & 36 \\
Century & 48 & 12 & 4 & -- & -- \\
Wingspan & 48 & 12 & 4 & -- & -- \\
Azul & 48 & 12 & 4 & -- & -- \\
Harmonies & 48 & 12 & 4 & -- & -- \\
Carcassonne & 48 & 12 & 4 & -- & -- \\
7 Wonders & 48 & 12 & 4 & -- & 36 \\
Burgundy & 48 & 12 & 4 & -- & -- \\
Brass: Birmingham & 48 & 12 & 4 & 72 & 36 \\
Ark Nova & 48 & 12 & 4 & -- & -- \\
\textbf{Total} & 480 & 120 & 40 & 144 & 108 \\
\bottomrule
\end{tabular*}

\end{table}

The original cohort contributes 144 three-player matches and the seven-game
expansion contributes 336. Adding the 252 supplementary matches gives 732
native matches. The separate 72-match bridge and 48-match within-Qwen-family
study are excluded from these totals.
The separate Random, Heuristic, and ISMCTS reference cohorts are
also excluded from these native-cohort totals.

PWS measures the fraction of opponents outranked, with half credit for ties;
the ten-game mean weights games equally. Elo adjusts for the observed
comparison graph, so its ordering need not equal unadjusted PWS ordering.
Table~\ref{tab:expanded-elo} provides all ten per-game Elo fits from the
primary cohort. Their intervals and PWS uncertainty are documented in
Appendices~\ref{app:completed-inference} and~\ref{app:score-intervals}.
\begin{table}[!htbp]
\centering\footnotesize
\caption{Three-player Elo point estimates for all ten native games,
rounded to integers. Each game is fitted separately over the same 12
configurations and centered at 1500. No average of per-game Elo is used.
IDs and game abbreviations follow Table~\ref{tab:performance-compact}.}
\label{tab:expanded-elo}
\setlength{\tabcolsep}{1.5pt}
\begin{tabular*}{\linewidth}{@{\extracolsep{\fill}}lr*{10}{r}@{}}
\toprule
Configuration & ID & SP & CE & WI & AZ & HA & CA & 7W & BU & BR & AN \\
\midrule
GPT-5.6-Sol / Codex & 1 & 1622 & 1686 & 1811 & 1609 & 1821 & 1634 & 1634 & 1699 & 2068 & 2143 \\
GPT-5.6-Sol / OpenCode & 2 & 1552 & 1908 & 1728 & 1629 & 1888 & 1789 & 1695 & 1884 & 1673 & 1902 \\
GPT-5.6-Terra / Codex & 3 & 1419 & 1746 & 1511 & 1385 & 1328 & 1385 & 1445 & 1435 & 1794 & 1752 \\
GPT-5.6-Luna / Codex & 4 & 1275 & 1105 & 1170 & 1414 & 1329 & 1327 & 1297 & 1413 & 1328 & 1432 \\
DSV4-Pro / DSH & 5 & 1567 & 1646 & 1603 & 1561 & 1561 & 1641 & 1620 & 1590 & 1591 & 1614 \\
DSV4-Pro / OpenCode & 6 & 1550 & 1613 & 1378 & 1478 & 1660 & 1477 & 1583 & 1598 & 1525 & 1605 \\
DSV4-Flash / DSH & 7 & 1500 & 1427 & 1468 & 1528 & 1358 & 1367 & 1512 & 1494 & 1359 & 1320 \\
Qwen3.8-Max / OpenCode & 8 & 1493 & 1656 & 1687 & 1459 & 1822 & 1571 & 1545 & 1413 & 1497 & 1759 \\
Qwen3.8-Max / Qwen Code & 9 & 1654 & 1787 & 1837 & 1734 & 2010 & 1924 & 1439 & 1756 & 1515 & 1349 \\
Qwen3.7-Plus / OpenCode & 10 & 1278 & 591 & 1194 & 1232 & 900 & 1225 & 1427 & 969 & 1291 & 845 \\
GLM-5.2 / OpenCode & 11 & 1587 & 1551 & 1328 & 1457 & 1378 & 1347 & 1539 & 1475 & 1310 & 1228 \\
Kimi-K2.7-Code / OpenCode & 12 & 1502 & 1284 & 1287 & 1512 & 945 & 1312 & 1264 & 1273 & 1051 & 1050 \\
\bottomrule
\end{tabular*}

\end{table}

Source reconciliation and database checks are described once in
Appendix~\ref{app:reproducibility}. Performance and resource comparisons
use the same 480-match primary cohort (Appendix~\ref{app:elo-cost}).

\subsection{Placement measurement for the primary cohort}
\label{app:pws-diagnostic}
For participant $i$ with rank $r_{it}$ in an $n_t$-player match $t$, define
\begin{equation}
u_{it}=\frac{1}{n_t-1}\sum_{j\ne i}
\left[\mathbb{1}(r_{it}<r_{jt})+
\tfrac12\mathbb{1}(r_{it}=r_{jt})\right].
\end{equation}
This pairwise placement score credits tied opponents by one half. For the
primary cohort, $n_t=3$ throughout. With $T_{ig}$ denoting participant $i$'s
12 appearances in game $g$, report
\begin{equation}
S_{ig}=\frac{1}{|T_{ig}|}\sum_{t\in T_{ig}}u_{it},
\qquad S_i=\frac{1}{10}\sum_{g\in\mathcal G}S_{ig}.
\label{eq:pws}
\end{equation}
The ten games receive equal weight. Supplementary two-/four-player matches,
non-LLM baselines, and matched harness studies are separate from this estimand.

Whole-match resampling within each game retains co-players and final rankings
together, recomputes cell means and equal-game averages, and refits Elo
separately. Pointwise percentile intervals condition on the evaluated games
and schedules.

\subsection{Primary-cohort uncertainty and ranking agreement}
\label{app:completed-inference}
The primary cohort contains 480 unique three-player matches,
1,440 participant appearances, and 120 participant--game cells. Each cell
has 12 appearances, with four at each seat. Every PWS cell is computed from
final ranks, with half credit for ties. The resulting 12 configuration
means sum to 600 percentage points within each game before rounding.

\paragraph{Sampling intervals.}
We draw 5,000 whole-match bootstrap samples separately within each of the
ten games, using seed 20260924. Players, opponents, and terminal outcomes
remain together. Each draw recomputes all game-specific PWS values,
equal-game means, and the between-group configuration contrasts in
Table~\ref{tab:configuration-concentration}. Games and group membership
remain fixed. One draw omitted a configuration in Burgundy and was redrawn.
The intervals are pointwise 95\% percentile intervals. Table~\ref{tab:completed-aggregate}
reports aggregate PWS.

\begin{table}[!htbp]
\centering\footnotesize
\caption{Ten-game equal-weight PWS and pointwise 95\% whole-match bootstrap
intervals. Each configuration has 120 primary-cohort appearances. These
estimates exclude non-LLM baselines and the matched-bridge cohort.}
\label{tab:completed-aggregate}
\begin{tabular*}{\linewidth}{@{\extracolsep{\fill}}lrr@{}}
\toprule
Configuration & Mean PWS (\%) & 95\% interval \\
\midrule
GPT-5.6-Sol / OpenCode & 75.6 & [69.7, 81.3] \\
GPT-5.6-Sol / Codex & 75.4 & [68.9, 81.7] \\
Qwen3.8-Max / Qwen Code & 70.2 & [64.5, 75.7] \\
DSV4-Pro / DSH & 60.4 & [53.6, 67.0] \\
Qwen3.8-Max / OpenCode & 58.3 & [51.9, 64.7] \\
DSV4-Pro / OpenCode & 54.2 & [47.3, 60.8] \\
GPT-5.6-Terra / Codex & 50.2 & [43.4, 57.0] \\
DSV4-Flash / DSH & 42.9 & [35.3, 50.3] \\
GLM-5.2 / OpenCode & 41.9 & [35.0, 48.8] \\
GPT-5.6-Luna / Codex & 29.6 & [23.8, 35.7] \\
Kimi-K2.7-Code / OpenCode & 25.2 & [19.6, 30.9] \\
Qwen3.7-Plus / OpenCode & 16.0 & [11.3, 21.2] \\
\bottomrule
\end{tabular*}

\end{table}

For Elo, each sample refits the same centered, penalized Bradley--Terry
objective used for the point estimate. Three-player pair weights are
$1/2$, the unnormalized loss uses penalty $0.02\sum_i\theta_i^2/2$,
and Elo equals $1500+400\theta_i/\log(10)$. Newton optimization converges
this objective. All bootstrap comparison graphs remain connected.
Independent fits reproduce all 24 aggregate Elo estimates to one decimal
place. For the original three-game cohort, equal-game/equal-player-count
PWS estimates agree to four decimal places.

\paragraph{Supplementary leave-one-game-out check.}
For each target game $g$, average each configuration's PWS over the other
nine games and compare that ranking with its PWS ranking in $g$ using
Kendall $\tau_b$, which handles ties. Pair-order accuracy excludes pairs
tied in the target game and gives half credit when the training scores tie.
No held-out outcomes enter the nine-game ranking. This analysis describes
how configuration rankings transfer across the ten games.

\begin{table}[!htbp]
\centering\footnotesize
\caption{Descriptive cross-game ranking agreement and within-game
split-half reliability. Game abbreviations match
Table~\ref{tab:performance-compact}. The final range describes variation
across random splits; it is not a confidence interval.}
\label{tab:completed-loeo}
\begin{tabular*}{\linewidth}{@{\extracolsep{\fill}}lrrrl@{}}
\toprule
Game & LOEO $\tau_b$ & Order acc. & Split-half mean & Split 5--95\% range \\
\midrule
SP & 0.49 & 75.4\% & 0.18 & [-0.07, 0.44] \\
CE & 0.72 & 86.2\% & 0.58 & [0.38, 0.78] \\
WI & 0.72 & 86.2\% & 0.55 & [0.33, 0.74] \\
AZ & 0.59 & 80.6\% & 0.27 & [0.00, 0.53] \\
HA & 0.71 & 85.9\% & 0.64 & [0.47, 0.80] \\
CA & 0.86 & 93.8\% & 0.49 & [0.27, 0.71] \\
7W & 0.58 & 80.3\% & 0.28 & [0.03, 0.54] \\
BU & 0.69 & 84.9\% & 0.49 & [0.28, 0.70] \\
BR & 0.69 & 85.5\% & 0.53 & [0.32, 0.73] \\
AN & 0.60 & 80.0\% & 0.69 & [0.51, 0.84] \\
\bottomrule
\end{tabular*}

\end{table}

Mean LOEO $\tau_b$ is 0.664 and equal-game pair-order accuracy is 83.9\%.
For each game's reliability control, 2,000 disjoint random splits divide
its 48 matches into two 24-match sets, conditioned on all configurations
appearing in both sets. Each split preserves complete matches but does
not force equal seat or opponent exposure within each half. Mean split-half
$\tau_b$ is 0.469 across games; Splendor is 0.176 and Ark Nova is 0.687.
The 5th--95th percentile split ranges appear in Table~\ref{tab:completed-loeo}.
LOEO uses full-game samples; the reliability control uses half-game samples.

Additional whole-match perturbations yield a 2.5th--97.5th percentile
range of 0.480--0.639 for mean LOEO $\tau_b$, below the plug-in estimate.
Rank discreteness and the extra sampling noise make this a sensitivity
distribution rather than a calibrated confidence interval. The descriptive
LOEO estimates indicate common overall strength alongside game-specific deviations.

\subsection{Strategic specialization: definitions and descriptive checks}
\label{app:specialization}
RQ1 uses the ten-game three-player PWS matrix and the corresponding
Elo ratings estimated from the same completed matches.
Figure~\ref{fig:elo-cost} uses those same game-specific ratings.
Rank each game's Elo over the same 12 configurations; assign average ranks
for ties when comparing PWS and Elo orderings. Table~\ref{tab:specialization-ranks}
shows the configurations discussed in the text. 

\begin{table}[!htbp]
\centering\footnotesize
\caption{Selected configurations' three-player Elo ranks among all 12
participants in each game; 1 is highest. Columns follow
Table~\ref{tab:performance-compact}.}
\label{tab:specialization-ranks}
\begin{tabular*}{\linewidth}{@{\extracolsep{\fill}}l*{10}{r}@{}}
\toprule
Configuration & SP & CE & WI & AZ & HA & CA & 7W & BU & BR & AN \\
\midrule
GPT-5.6-Sol / Codex & 2 & 4 & 2 & 3 & 4 & 4 & 2 & 3 & 1 & 1 \\
GPT-5.6-Sol / OpenCode & 5 & 1 & 3 & 2 & 2 & 2 & 1 & 1 & 3 & 2 \\
GPT-5.6-Terra / Codex & 10 & 3 & 6 & 11 & 10 & 7 & 8 & 8 & 2 & 4 \\
Qwen3.8-Max / Qwen Code & 1 & 2 & 1 & 1 & 1 & 1 & 9 & 2 & 6 & 8 \\
Qwen3.8-Max / OpenCode & 9 & 5 & 4 & 8 & 3 & 5 & 5 & 10 & 7 & 3 \\
DSV4-Pro / DSH & 4 & 6 & 5 & 4 & 6 & 3 & 3 & 5 & 4 & 5 \\
\bottomrule
\end{tabular*}

\end{table}

\paragraph{Group definitions and interpretation.}
The public-information spatial group combines spatial planning = 2 and
hidden information = 0 in Appendix~\ref{app:game-profiles}: Azul, Harmonies,
Carcassonne, and Burgundy. Games with spatial planning = 2 additionally
include Brass and Ark Nova; the remaining games are Splendor, Century,
Wingspan, and 7 Wonders. Games with hidden information = 2 comprise
Wingspan, 7 Wonders, Brass, and Ark Nova. Hidden-information annotations
exclude uncertainty about future draws, so public holdings do not imply
deterministic gameplay.
Groups overlap and contain different combinations of other mechanisms.
The Century/Brass/Ark Nova and Azul/Harmonies/Carcassonne triads describe
GPT-5.6-Terra's performance profile across these games.

\paragraph{Concentration of configuration advantages.}
For configurations $A$ and $B$ of the same model, define
$\delta_g=100(\widehat{\mathrm{PWS}}_{A,g}-\widehat{\mathrm{PWS}}_{B,g})$.
Let $G$ be the four public-information spatial games and $H$ the other
six: Splendor, Century, Wingspan, 7 Wonders, Brass, and Ark Nova. Average
games equally within each group and report
\begin{equation}
\overline\Delta_G=\frac{1}{|G|}\sum_{g\in G}\delta_g,
\qquad
\overline\Delta_H=\frac{1}{|H|}\sum_{g\in H}\delta_g,
\qquad
\Gamma=\overline\Delta_G-\overline\Delta_H.
\end{equation}
Table~\ref{tab:configuration-concentration} includes all three nominal models
with two native configurations: Qwen3.8-Max (Qwen Code minus OpenCode),
GPT-5.6-Sol (Codex minus OpenCode), and DSV4-Pro (DSH minus OpenCode).
$\Gamma$ distinguishes a group's configuration advantage from the average
advantage elsewhere. A uniform shift in $\delta_g$ across all ten games
cancels in $\Gamma$. The reference group combines several game designs,
and each model pair compares its own two harnesses. These are descriptive
comparisons of complete configurations under their native opponent samples.

\paragraph{Game-composition sensitivity.}
Omit one game at a time from whichever group contains it and recompute
$\Gamma$ over the remaining games, preserving their group assignments.
Across all ten deletions, the ranges are 23.0--36.1 pp for Qwen3.8-Max,
$-19.7$--$-9.7$ pp for GPT-5.6-Sol, and $-5.6$--$+1.9$ pp for DSV4-Pro.
These are descriptive ranges over game subsets, not confidence intervals.

\begin{table}[!htbp]
\centering\footnotesize
\caption{Additional game-deletion diagnostics for Qwen3.8-Max: Qwen Code minus
OpenCode PWS. Public spatial means spatial planning = 2 and hidden
information = 0; hidden information = 2 denotes central private holdings.
These within-group means differ from the between-group
contrast in Table~\ref{tab:configuration-concentration}.}
\label{tab:specialization-sensitivity}
\begin{tabular*}{\linewidth}{@{\extracolsep{\fill}}lr@{}}
\toprule
Descriptive check (Qwen3.8-Max throughout) & PWS difference (pp) \\
\midrule
Public spatial: Qwen Code minus OpenCode & $+29.7$ \\
Same contrast, omit Azul & $+28.5$ \\
Same contrast, omit Harmonies & $+36.1$ \\
Same contrast, omit Burgundy & $+25.7$ \\
Same contrast, omit Carcassonne & $+28.5$ \\
HI=2: Qwen Code minus OpenCode & $-8.3$ \\
HI=2 contrast, omit Ark Nova & $+0.0$ \\
\bottomrule
\end{tabular*}

\end{table}

Table~\ref{tab:specialization-sensitivity} retains the within-group diagnostics.
Each public-information spatial game has a positive Qwen Code--OpenCode
PWS contrast, so every one-game deletion preserves a positive mean.
For games with hidden information = 2, the negative mean disappears when
Ark Nova is removed, showing that this game's result drives the group average.

\paragraph{Metric agreement and uncertainty.}
Within-game Spearman agreement between Elo and PWS ranks ranges from
0.975 to 0.998 across games. The two rankings derive from the same
outcomes; their agreement checks sensitivity to the choice of ranking metric.

Pointwise 95\% intervals for $\Gamma$ use 5,000 whole-match bootstrap
resamples within each game (Appendix~\ref{app:completed-inference}).
Each draw retains all co-players, recomputes configuration differences,
and averages games within the fixed groups.
The public-spatial minus other-games contrast is 29.7 pp [11.9, 46.9] for
Qwen3.8-Max, $-14.4$ pp [$-32.6$, 4.2] for GPT-5.6-Sol, and $-1.7$ pp
[$-21.8$, 18.1] for DSV4-Pro. Sampling intervals describe variation across
matches; game deletion measures sensitivity to the selected games.

\subsection{Complementary native-system contrasts}
\label{app:native-contrasts}
\begin{figure}[!htbp]
\centering
\includegraphics[width=\linewidth]{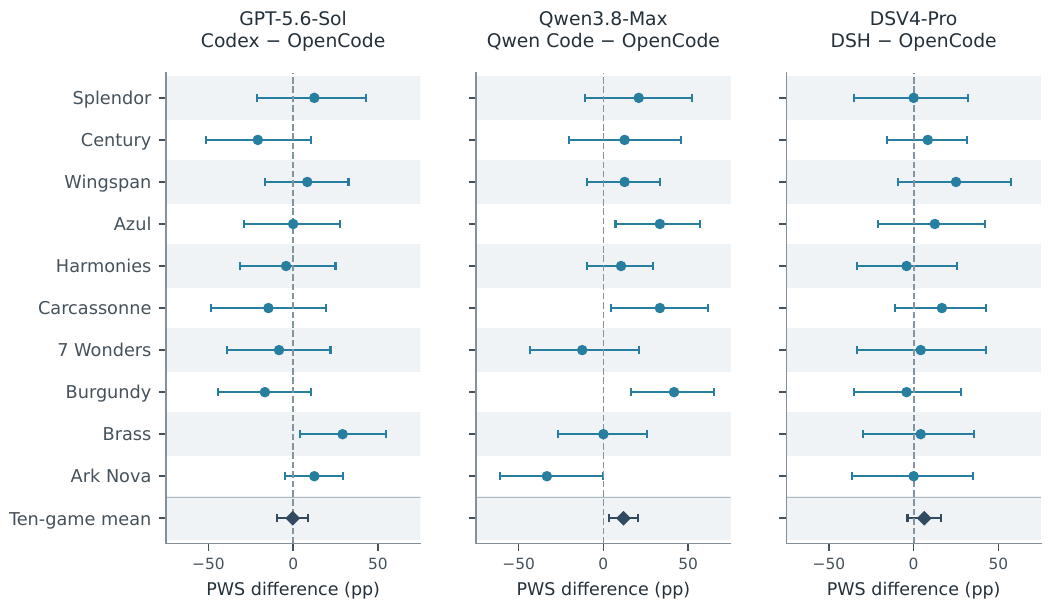}
\caption{Same-model PWS contrasts across all ten native games in the
480-match three-player cohort. Positive values favor the first named
harness; diamonds show equal-game means. Pointwise 95\% intervals reuse
the joint whole-match bootstrap draws, preserving covariance between
configurations within a game. These descriptive native-system contrasts
use different opponent samples and are separate from the matched studies.}
\label{fig:contrasts}
\end{figure}

The ten-game mean GPT-5.6-Sol Codex--OpenCode difference is $-0.2$ pp
[$-9.3$, 8.6], despite game-level differences ranging from $-20.8$ in
Century to $+29.2$ in Brass. Qwen3.8-Max / Qwen Code minus OpenCode averages
$+11.9$ pp [3.2, 20.5], ranging from $-33.3$ in Ark Nova to $+41.7$ in
Burgundy. DSV4-Pro / DSH minus OpenCode averages $+6.3$ pp [$-3.6$, 16.0].
These full-game profiles complement the group concentrations in
Table~\ref{tab:configuration-concentration}.

\subsection{Ten-game performance with uncertainty}
\label{app:score-intervals}
\begin{figure}[!htbp]
\centering
\includegraphics[width=\linewidth]{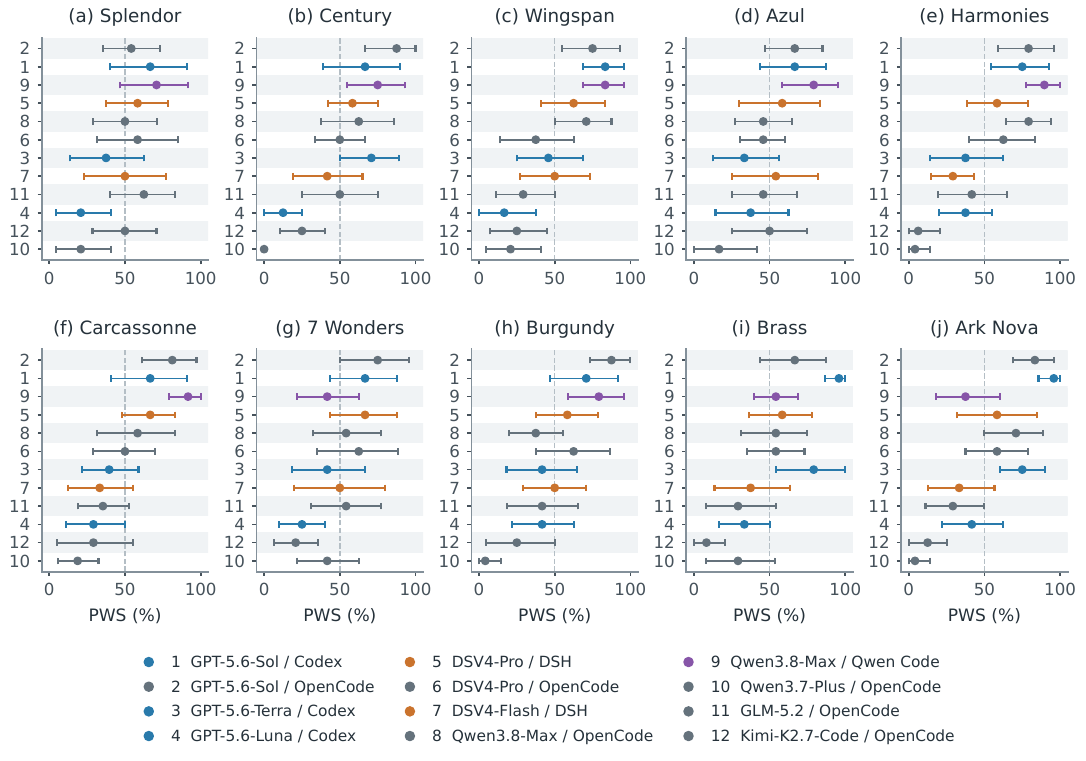}
\caption{Per-game PWS with pointwise 95\% whole-match bootstrap intervals
for all ten native games. IDs match Table~\ref{tab:performance-compact};
every panel orders configurations by their ten-game mean PWS. Each panel
contains 48 three-player matches and 12 appearances per configuration.
The dashed reference is 50\% PWS.}
\label{fig:rankings}
\end{figure}

Figure~\ref{fig:rankings} and Table~\ref{tab:full-scores} expand the main
performance matrix to show uncertainty for all 120 configuration--game
cells in the 480-match three-player cohort. Each cell has 12 appearances;
the ten-game mean and its interval appear in Table~\ref{tab:completed-aggregate}.

\begin{table}[!htbp]
\centering\scriptsize
\caption{Ten-game PWS (\%) and pointwise 95\% whole-match bootstrap intervals.
Panels cover all 120 primary cells; IDs and game abbreviations match
Table~\ref{tab:performance-compact}. Each cell uses 12 appearances.
Intervals are computed jointly within each game's sampled matches and
are not simultaneous bounds across cells.}
\label{tab:full-scores}
\setlength{\tabcolsep}{3pt}
\noindent\textbf{(a) Splendor through Harmonies}\par\smallskip
\begin{tabular*}{\linewidth}{@{\extracolsep{\fill}}r*{5}{r}@{}}
\toprule
ID & SP & CE & WI & AZ & HA \\
\midrule
1 & 66.7 [40.0, 90.9] & 66.7 [38.9, 90.0] & 83.3 [68.8, 95.8] & 66.7 [43.8, 87.5] & 75.0 [54.5, 92.9] \\
2 & 54.2 [35.7, 73.1] & 87.5 [66.7, 100.0] & 75.0 [55.0, 92.9] & 66.7 [47.2, 85.0] & 79.2 [59.1, 95.8] \\
3 & 37.5 [13.6, 62.5] & 70.8 [50.0, 88.9] & 45.8 [25.0, 68.8] & 33.3 [12.5, 56.2] & 37.5 [14.3, 62.5] \\
4 & 20.8 [4.2, 40.9] & 12.5 [0.0, 25.0] & 16.7 [0.0, 37.5] & 37.5 [14.3, 62.5] & 37.5 [20.0, 55.0] \\
5 & 58.3 [37.5, 78.1] & 58.3 [42.3, 75.0] & 62.5 [40.9, 83.3] & 58.3 [30.0, 83.3] & 58.3 [38.2, 78.6] \\
6 & 58.3 [31.8, 85.0] & 50.0 [33.3, 66.7] & 37.5 [13.6, 62.5] & 45.8 [30.6, 60.0] & 62.5 [40.0, 83.3] \\
7 & 50.0 [22.7, 76.9] & 41.7 [19.2, 65.0] & 50.0 [26.9, 73.5] & 54.2 [25.0, 81.8] & 29.2 [15.0, 42.9] \\
8 & 50.0 [29.2, 71.4] & 62.5 [37.5, 85.7] & 70.8 [50.0, 87.5] & 45.8 [26.9, 65.0] & 79.2 [64.3, 93.8] \\
9 & 70.8 [46.4, 91.7] & 75.0 [55.0, 92.9] & 83.3 [68.8, 95.8] & 79.2 [58.3, 95.5] & 89.6 [77.5, 100.0] \\
10 & 20.8 [4.2, 40.9] & 0.0 [0.0, 0.0] & 20.8 [4.5, 40.9] & 16.7 [0.0, 41.7] & 4.2 [0.0, 14.3] \\
11 & 62.5 [40.0, 83.3] & 50.0 [25.0, 75.0] & 29.2 [11.1, 50.0] & 45.8 [25.0, 68.2] & 41.7 [19.2, 65.0] \\
12 & 50.0 [28.6, 70.8] & 25.0 [10.7, 40.0] & 25.0 [7.1, 45.0] & 50.0 [25.0, 75.0] & 6.2 [0.0, 20.5] \\
\bottomrule
\end{tabular*}

\par\medskip
\noindent\textbf{(b) Carcassonne through Ark Nova}\par\smallskip
\begin{tabular*}{\linewidth}{@{\extracolsep{\fill}}r*{5}{r}@{}}
\toprule
ID & CA & 7W & BU & BR & AN \\
\midrule
1 & 66.7 [40.9, 90.9] & 66.7 [43.8, 87.5] & 70.8 [47.1, 91.7] & 95.8 [86.4, 100.0] & 95.8 [85.7, 100.0] \\
2 & 81.2 [61.1, 97.2] & 75.0 [50.0, 95.8] & 87.5 [73.1, 100.0] & 66.7 [43.8, 87.5] & 83.3 [68.8, 96.2] \\
3 & 39.6 [21.4, 58.9] & 41.7 [18.2, 66.7] & 41.7 [18.2, 65.0] & 79.2 [54.5, 100.0] & 75.0 [60.0, 90.0] \\
4 & 29.2 [11.1, 50.0] & 25.0 [10.0, 40.0] & 41.7 [21.9, 62.5] & 33.3 [16.7, 50.0] & 41.7 [22.2, 62.5] \\
5 & 66.7 [47.8, 83.3] & 66.7 [43.8, 87.5] & 58.3 [37.5, 78.6] & 58.3 [36.4, 77.8] & 58.3 [31.8, 84.6] \\
6 & 50.0 [29.2, 70.0] & 62.5 [35.0, 88.5] & 62.5 [37.5, 86.4] & 54.2 [35.3, 73.1] & 58.3 [37.5, 78.6] \\
7 & 33.3 [12.5, 55.6] & 50.0 [20.0, 80.0] & 50.0 [29.2, 70.8] & 37.5 [13.6, 63.6] & 33.3 [12.5, 56.7] \\
8 & 58.3 [31.8, 83.3] & 54.2 [32.1, 76.9] & 37.5 [20.0, 55.6] & 54.2 [31.2, 75.0] & 70.8 [50.0, 88.9] \\
9 & 91.7 [79.2, 100.0] & 41.7 [21.9, 62.5] & 79.2 [59.1, 95.8] & 54.2 [40.0, 68.8] & 37.5 [17.9, 60.0] \\
10 & 18.8 [5.6, 32.5] & 41.7 [21.9, 62.5] & 4.2 [0.0, 14.3] & 29.2 [8.3, 53.6] & 4.2 [0.0, 14.3] \\
11 & 35.4 [18.8, 52.8] & 54.2 [31.2, 77.3] & 41.7 [18.8, 65.4] & 29.2 [8.3, 54.2] & 29.2 [11.1, 50.0] \\
12 & 29.2 [5.0, 55.6] & 20.8 [6.7, 35.7] & 25.0 [4.5, 50.0] & 8.3 [0.0, 20.6] & 12.5 [0.0, 25.0] \\
\bottomrule
\end{tabular*}

\par\medskip

\end{table}

\subsection{Cohort provenance and reproducibility}
\label{app:reproducibility}
The native study includes 396 matches in the original cohort and 336 in
the expansion. We reconcile individual match results with the combined
outcome records and check each match's game, player count, roster, seats,
and seed against the schedule. All 732 final game records pass integrity
checks and agree on match identity, game, rules version, seed, and terminal
status. Their 528 stored terminal ranking events agree with the reported
ranks; Splendor and Century rankings agree between individual and combined
outcome records. Checksums verify all 672 result and game-state records
in the expansion.

Deterministic analyses compute all 120 primary PWS cells from final ranks,
fit per-game Elo, and generate 5,000 whole-match bootstrap replicates and
the supplementary ranking diagnostics. Independent calculations reproduce
aggregate PWS and Elo estimates to rounding precision.
The native configuration contrasts use the same whole-match draws.
All 120 per-game and 12 equal-game PWS intervals reconcile exactly.

DECE supports replay and resume through persisted game states and execution
records. The analysis associates each scheduled match with its final
terminal outcome. Native telemetry covers all 540
Codex appearances across the two source cohorts, including 360 in the
primary three-player cohort. Native session records also cover six
GPT-5.6-Sol / OpenCode appearances.
The resource accounting and verification procedures are described
in Appendix~\ref{app:elo-cost}.

Matched schedules and separate baseline records reproduce PWS
for all 180 non-LLM reference matches
(Appendix~\ref{app:nonllm-baselines}).
The native performance, horizon, execution, and same-model contrast figures
use the ten-game three-player cohort. Additional player-count analyses remain
separate; their coverage is explicitly limited to observed strata.

The within-Qwen-family study includes 48 complete outcomes and 24 matched
harness pairs, with independent outcome verification
(Appendix~\ref{app:qwen-harness-pilot}). Its final game records pass integrity
and terminal-identity checks, and all focal PWS values agree with exact
rank-based arithmetic. Figure~\ref{fig:matched-bridge} and the within-Qwen-family
statistics are computed directly from the matched outcomes.

\subsection{Non-LLM reference baselines}
\label{app:nonllm-baselines}
\label{app:random-baseline}
Random samples uniformly from legal-action entries.
Heuristic uses fixed, game-specific action scores with an $\epsilon$-greedy
policy ($\epsilon=0.1$; \citealp[Section~2.2]{sutton2018reinforcement}), without tree search.
ISMCTS uses these scores for rollouts in single-observer information set
Monte Carlo tree search \citep{cowling2012ismcts}.
It accesses the native simulator with at most 1,024 simulations and 18 search
steps per decision, under a 30-second wall-time target.

The completed study contains 180 three-player matches: 60 per baseline and
six per game, each against
two LLM opponents. Opponents, seeds, rules, and focal seats match across the
three baseline schedules. Each LLM configuration appears once per game;
each baseline occupies every seat twice.
These cohorts are separate from the main PWS, Elo, cross-game, and resource analyses.

The ten-game mean PWS is 0.8\% for Random, 10.0\% for Heuristic,
and 44.6\% for ISMCTS.

\section{Elo and Resource-Efficiency Protocol}
\label{app:elo-cost}

\subsection{Cohort-aligned recorded resource profiles}
Figures~\ref{fig:elo-cost} and~\ref{fig:resource-overview} use the same 480
three-player matches as the primary performance table: 48 per game, 12
appearances per configuration/game, and 1,440 appearances overall. Resource
records are joined to the audited outcome ledger by final match ID, seat,
and participant identity. No two- or four-player cost is included, and
non-LLM baselines and the matched-bridge cohort are excluded.

The estimand is \emph{recorded consumption associated with a final successful
match ID}, including retained native responses, recorded controller usage,
and observable continuations. Replaced match IDs are outside this estimand.

\subsection{Native execution demand and usage coverage}
\label{app:telemetry}
\begin{figure}[!htbp]
\centering
\includegraphics[width=\linewidth]{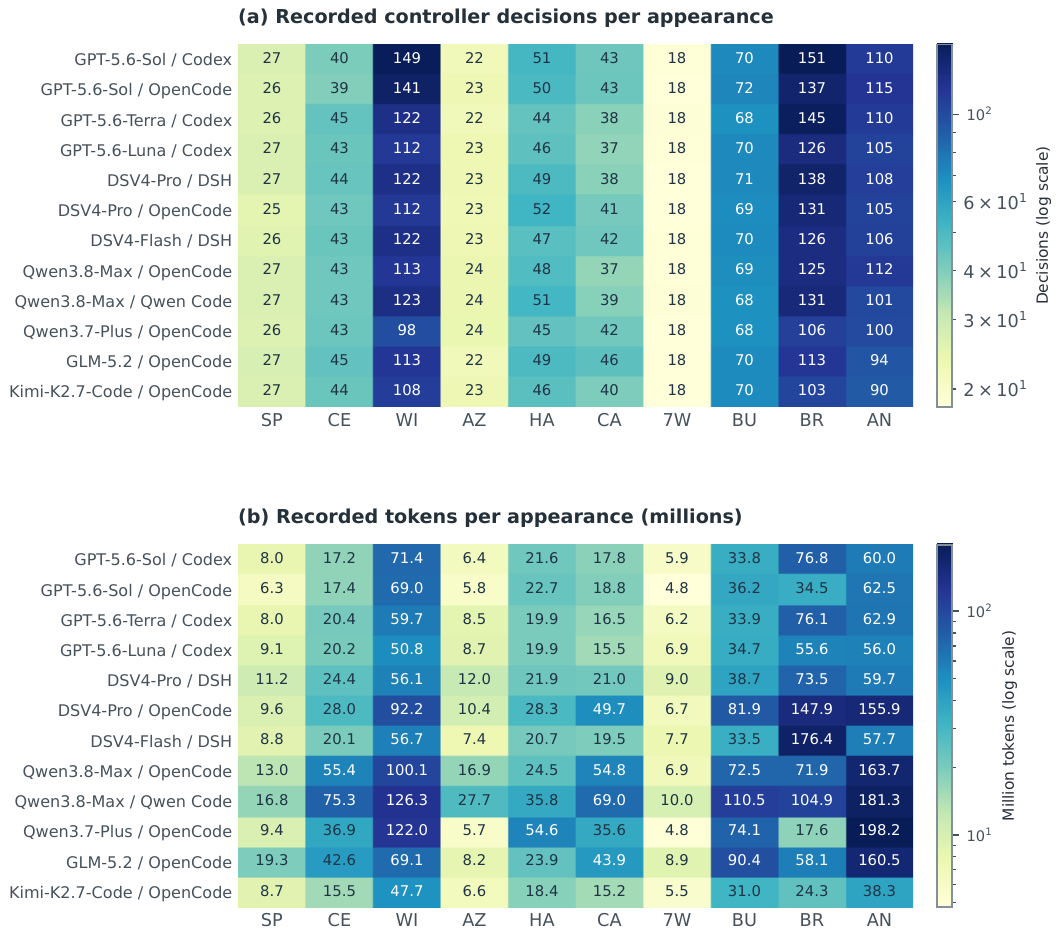}
\caption{Execution demand and recorded token volume across the ten-game
primary cohort. Cells are means over 12 appearances: (a) saved controller
decisions; (b) normalized input plus output tokens, in millions.
Each panel has its own logarithmic color scale; printed values use the
original units. Game abbreviations follow Table~\ref{tab:performance-compact}.}
\label{fig:operations}
\end{figure}

Figure~\ref{fig:operations} joins recorded controller decisions and normalized
token use to all 1,440 participant appearances in the same 480 matches.
Table~\ref{tab:telemetry} reports usage coverage from controller summaries
and from the combined controller and native session records.
\begin{table}[!htbp]
\centering\footnotesize
\caption{Ten-game execution and usage coverage, 120 appearances per
configuration. Decisions are totals from saved controller records.
Summary coverage counts appearances with numeric token totals in controller
summaries; combined coverage includes usage from native session records.
For Codex, native session logs provide token
usage for all 360 primary-cohort appearances (120 per configuration); the
controller summaries omit these totals.}
\label{tab:telemetry}
\setlength{\tabcolsep}{3pt}
\begin{tabular*}{\linewidth}{@{\extracolsep{\fill}}lrrrr@{}}
\toprule
 & & & \multicolumn{2}{c}{Token-usage coverage} \\
\cmidrule(lr){4-5}
Configuration & ID & Decisions & Summary coverage & Combined coverage \\
\midrule
GPT-5.6-Sol / Codex & 1 & 8,176 & 0/120 & 120/120 \\
GPT-5.6-Sol / OpenCode & 2 & 7,975 & 114/120 & 120/120 \\
GPT-5.6-Terra / Codex & 3 & 7,673 & 0/120 & 120/120 \\
GPT-5.6-Luna / Codex & 4 & 7,303 & 0/120 & 120/120 \\
DSV4-Pro / DSH & 5 & 7,656 & 114/120 & 120/120 \\
DSV4-Pro / OpenCode & 6 & 7,430 & 117/120 & 120/120 \\
DSV4-Flash / DSH & 7 & 7,473 & 116/120 & 120/120 \\
Qwen3.8-Max / OpenCode & 8 & 7,379 & 115/120 & 120/120 \\
Qwen3.8-Max / Qwen Code & 9 & 7,483 & 113/120 & 120/120 \\
Qwen3.7-Plus / OpenCode & 10 & 6,850 & 118/120 & 120/120 \\
GLM-5.2 / OpenCode & 11 & 7,180 & 119/120 & 120/120 \\
Kimi-K2.7-Code / OpenCode & 12 & 6,832 & 117/120 & 120/120 \\
\bottomrule
\end{tabular*}

\end{table}

Controller decisions, engine decisions, and native model responses are
separate units. A saved controller decision may contain several tool calls,
and native usage can be available even when the controller summary lacks a
token total. Usage accounting combines verified Codex native records, incremental
controller usage where available, and Qwen Code native session totals or
deduplicated native API responses.
Input totals include cached input; output totals include reasoning when
reported as a subset. All 1,440 appearances have quantified usage.
The accounting scope, price
scenario, and resource uncertainty are defined in Appendix~\ref{app:elo-cost}.

\subsection{Accounting adapters and verification}
\paragraph{Codex.}
Original-cohort usage is computed from cumulative native session counters.
Duplicate snapshots contribute zero; increments and resets are checked
against last-response usage before summation. The expansion instead sums
unique native responses and checks them against source-log entries and
seat totals. Together these sources cover all 360 primary Codex appearances.
Native responses and controller decisions remain distinct units.

\paragraph{Qwen Code.}
We use native session totals for the 36 original three-player
appearances and native API responses for the 84 expansion appearances.
Expansion records are deduplicated by match, seat, session, and response ID,
including context-compression requests. Incremental controller usage is
not treated as a cumulative session total.

\paragraph{OpenCode and DSH.}
We use incremental controller usage for the original three-game cohort.
For the seven-game expansion, we reconcile all OpenCode and DSH appearances
against native responses, including retained continuations and delegated calls.
Copied session histories are counted once. Native session records for six
GPT-5.6-Sol / OpenCode appearances agree with the recorded totals and are
included once.

\subsection{Frozen price scenario and token categories}
Input includes cached input as a subset; output includes reasoning when
reported as a subset. For an appearance with input $I$, cached input $H$,
and output $O$, the recorded API-equivalent scenario cost is
\begin{equation}
C=10^{-6}\{p_{\rm in}(I-H)+p_{\rm cache}H+p_{\rm out}O\}.
\label{eq:token-cost}
\end{equation}
All categories are checked for nonnegativity and subset consistency.
Input-plus-output token volume is $I+O$: cached and reasoning tokens are
not added again. Token volume is reported in each model's native tokenizer.

\begin{table}[!htbp]
\centering\scriptsize
\caption{Fixed price scenario, USD per million tokens, dated 28 August
2026 and converted at 6.7811 CNY per USD. Rates are fixed across
the resource analysis.}
\label{tab:price-scenario}
\setlength{\tabcolsep}{5pt}
\begin{tabular}{l rrr}
\toprule
Nominal model & Uncached input & Cached input & Output \\
\midrule
GPT-5.6-Sol & 4.0000 & 0.4000 & 20.0000 \\
GPT-5.6-Terra & 2.0000 & 0.2000 & 12.0000 \\
GPT-5.6-Luna & 0.2000 & 0.0200 & 1.2000 \\
DSV4-Flash & 0.4424 & 0.0442 & 1.3272 \\
DSV4-Pro & 1.3272 & 0.1327 & 3.9817 \\
GLM-5.2 & 1.1797 & 0.2949 & 4.1291 \\
Kimi-K2.7-Code & 0.9585 & 0.1917 & 3.9817 \\
Qwen3.7-Plus & 0.2359 & 0.0472 & 0.9438 \\
Qwen3.8-Max & 1.7696 & 0.2212 & 5.3089 \\
\bottomrule
\end{tabular}

\end{table}

The scenario uses DeepSeek busy-hour rates and the Qwen3.7-Plus
promotional input tier. Context-length and cache-write surcharges and
non-token charges are outside this scenario. Each configuration uses the
same nominal-model API rates across harnesses. Rates are applied before aggregation,
without rounding token means to millions.

\begin{table}[!htbp]
\centering\scriptsize
\caption{Recorded API-equivalent USD per participant appearance, twelve
appearances in every cell. Games follow Table~\ref{tab:performance-compact}.
Costs use Equation~\ref{eq:token-cost}.}
\label{tab:expanded-cost}
\begin{tabular*}{\linewidth}{@{\extracolsep{\fill}}lr*{10}{r}@{}}
\toprule
Configuration & ID & SP & CE & WI & AZ & HA & CA & 7W & BU & BR & AN \\
\midrule
GPT-5.6-Sol / Codex & 1 & 4.95 & 10.29 & 38.90 & 3.73 & 12.06 & 12.71 & 3.84 & 19.19 & 40.53 & 37.20 \\
GPT-5.6-Sol / OpenCode & 2 & 3.87 & 9.64 & 34.18 & 3.33 & 11.74 & 11.93 & 2.85 & 19.89 & 16.83 & 34.14 \\
GPT-5.6-Terra / Codex & 3 & 2.31 & 6.26 & 17.29 & 2.37 & 5.98 & 6.40 & 2.01 & 10.89 & 25.07 & 20.63 \\
GPT-5.6-Luna / Codex & 4 & 0.42 & 0.61 & 1.48 & 0.25 & 0.63 & 0.62 & 0.26 & 1.12 & 2.00 & 1.85 \\
DSV4-Pro / DSH & 5 & 2.53 & 4.92 & 9.19 & 2.82 & 4.20 & 5.81 & 2.38 & 7.25 & 13.69 & 11.36 \\
DSV4-Pro / OpenCode & 6 & 1.79 & 4.74 & 13.64 & 2.18 & 4.60 & 8.57 & 1.71 & 12.42 & 22.07 & 23.50 \\
DSV4-Flash / DSH & 7 & 0.61 & 2.18 & 5.11 & 1.05 & 1.81 & 2.91 & 0.69 & 3.77 & 9.09 & 6.13 \\
Qwen3.8-Max / OpenCode & 8 & 3.74 & 15.36 & 24.16 & 5.39 & 6.70 & 16.21 & 2.27 & 18.42 & 17.53 & 40.02 \\
Qwen3.8-Max / Qwen Code & 9 & 8.80 & 43.59 & 89.25 & 16.27 & 25.09 & 41.71 & 5.42 & 69.31 & 67.84 & 112.61 \\
Qwen3.7-Plus / OpenCode & 10 & 0.66 & 2.34 & 6.80 & 0.45 & 3.10 & 2.53 & 0.36 & 4.39 & 1.08 & 11.51 \\
GLM-5.2 / OpenCode & 11 & 6.90 & 14.00 & 21.90 & 3.03 & 8.20 & 15.26 & 3.27 & 29.00 & 18.95 & 51.55 \\
Kimi-K2.7-Code / OpenCode & 12 & 2.08 & 3.61 & 10.70 & 1.58 & 4.28 & 4.26 & 1.40 & 7.16 & 5.07 & 9.50 \\
\bottomrule
\end{tabular*}

\end{table}

\begin{table}[!htbp]
\centering\footnotesize
\caption{Equal-game resource overview. Each configuration contributes 120
three-player appearances across ten games. Cost intervals are pointwise
95\% whole-match percentile intervals conditional on the accounting and
price schedule. Tokens include cached input and reasoning exactly once.}
\label{tab:aggregate-resource}
\begin{tabular*}{\linewidth}{@{\extracolsep{\fill}}lrrr@{}}
\toprule
Configuration & PWS (\%) & USD [95\% interval] & Tokens (M) \\
\midrule
GPT-5.6-Sol / Codex & 75.4 & 18.34 [17.83, 18.89] & 31.89 \\
GPT-5.6-Sol / OpenCode & 75.6 & 14.84 [14.38, 15.35] & 27.79 \\
GPT-5.6-Terra / Codex & 50.2 & 9.92 [9.63, 10.20] & 31.21 \\
GPT-5.6-Luna / Codex & 29.6 & 0.92 [0.89, 0.96] & 27.74 \\
DSV4-Pro / DSH & 60.4 & 6.42 [6.21, 6.62] & 32.76 \\
DSV4-Pro / OpenCode & 54.2 & 9.52 [8.81, 10.28] & 61.06 \\
DSV4-Flash / DSH & 42.9 & 3.34 [3.15, 3.50] & 40.84 \\
Qwen3.8-Max / OpenCode & 58.3 & 14.98 [13.99, 16.04] & 57.95 \\
Qwen3.8-Max / Qwen Code & 70.2 & 47.99 [45.01, 51.09] & 75.77 \\
Qwen3.7-Plus / OpenCode & 16.0 & 3.32 [3.06, 3.60] & 55.91 \\
GLM-5.2 / OpenCode & 41.9 & 17.21 [16.17, 18.28] & 52.48 \\
Kimi-K2.7-Code / OpenCode & 25.2 & 4.96 [4.69, 5.30] & 21.11 \\
\bottomrule
\end{tabular*}

\end{table}

\subsection{Ratings, shared resampling, and descriptive frontiers}
Elo uses the same Bradley--Terry fits as the performance analysis
(Appendix~\ref{app:completed-inference}). Every unordered player pair
contributes a comparison with half credit for ties. The weighted
cross-entropy objective has penalty $0.02\sum_i\theta_i^2/2$ and
$\sum_i\theta_i=0$; ratings are $1500+400\theta_i/\log(10)$.
Newton optimization converges this batch objective. Pair weights are
$1/(n-1)=1/2$ for the three-player figure; mixed-player-count fits
use $1/[|\mathcal N_g|(n-1)]$ to weight strata equally. There is no
sequential $K$-factor update.

\begin{figure}[!htbp]
\centering
\includegraphics[width=\linewidth]{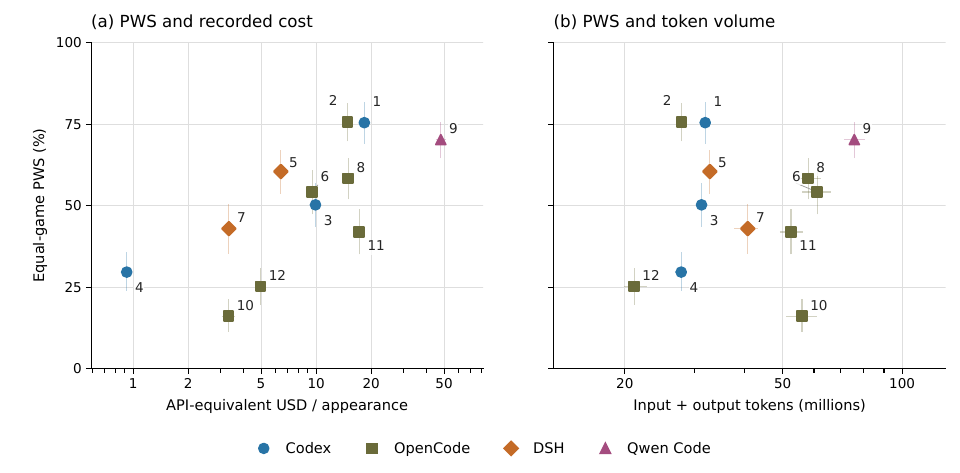}
\caption{Aggregate performance and resource use on the same 480 three-player
matches as Figure~\ref{fig:elo-cost}. Each point averages ten games equally,
with 120 appearances per configuration. Logarithmic horizontal axes show
mean API-equivalent USD per appearance and input-plus-output tokens.
Whiskers are pointwise 95\% whole-match intervals; IDs match
Table~\ref{tab:performance-compact}.}
\label{fig:resource-overview}
\end{figure}

The same 5,000 whole-match draws used for performance refit Elo and
recompute mean cost and token volume within each game. Opponents, outcomes,
and recorded costs stay together. Figure~\ref{fig:resource-overview} averages
game-level PWS and resource means equally, not separately centered Elo ratings.
Whiskers are marginal pointwise 95\% percentile intervals conditional on
the tested games, recorded usage, and fixed prices.

A tested configuration is nondominated if no other has at least as high
Elo and at most as much recorded cost, with one strict inequality. For
mean-cost budget $B$, the staircase in Figure~\ref{fig:elo-cost} is
\begin{equation}
V_g(B)=\max_{i:\,C_{ig}\leq B} E_{ig},
\end{equation}
where $C_{ig}$ is mean recorded cost and $E_{ig}$ is game-specific Elo.
The envelope is defined where at least one configuration is affordable.
Its steps select among tested configurations; $B$ constrains average cost,
not the cost of every individual match. Elo/cost ratios are not used
because Elo's origin is arbitrary.

\paragraph{Frontier coverage and sampling stability.}
For each configuration we count the games in whose point-estimate frontier
it appears, then recompute nondominance in every joint whole-match resample.
Table~\ref{tab:frontier-stability} reports these per-game resampling
frequencies. Frontier coverage includes both low-cost and high-performance
options. Frequencies describe sampling stability over the tested games
and price schedule.

\begin{table}[!htbp]
\centering\scriptsize
\setlength{\tabcolsep}{3pt}
\caption{Pareto-frontier coverage and bootstrap membership frequency
(\%). Games and configuration IDs follow Table~\ref{tab:performance-compact}.
Coverage counts point-estimate frontiers out of ten games; each frequency
uses 5,000 joint whole-match draws. Bold entries mark membership in the
point-estimate frontier.}
\label{tab:frontier-stability}
\begin{tabular}{lrr*{10}{r}}
\toprule
Configuration & ID & Games & Sp & Ce & Wi & Az & Ha & Ca & 7W & Bu & Br & AN\\
\midrule
GPT-5.6-Sol / Codex & 1 & 5/10 & \textbf{44.3} & 12.2 & \textbf{63.6} & 29.6 & 26.3 & 19.8 & 18.3 & \textbf{51.0} & \textbf{90.3} & \textbf{94.0}\\
GPT-5.6-Sol / OpenCode & 2 & 9/10 & 22.4 & \textbf{83.8} & \textbf{54.5} & \textbf{51.7} & \textbf{64.5} & \textbf{88.7} & \textbf{55.5} & \textbf{90.7} & \textbf{69.8} & \textbf{83.5}\\
GPT-5.6-Terra / Codex & 3 & 3/10 & 4.7 & \textbf{81.2} & 16.0 & 3.7 & 0.7 & 2.5 & 7.7 & 7.2 & \textbf{74.5} & \textbf{85.6}\\
GPT-5.6-Luna / Codex & 4 & 10/10 & \textbf{99.8} & \textbf{100.0} & \textbf{100.0} & \textbf{100.0} & \textbf{100.0} & \textbf{100.0} & \textbf{100.0} & \textbf{100.0} & \textbf{60.4} & \textbf{100.0}\\
DSV4-Pro / DSH & 5 & 10/10 & \textbf{37.0} & \textbf{71.4} & \textbf{81.5} & \textbf{42.6} & \textbf{71.5} & \textbf{98.7} & \textbf{49.6} & \textbf{76.5} & \textbf{93.1} & \textbf{92.0}\\
DSV4-Pro / OpenCode & 6 & 5/10 & \textbf{63.0} & \textbf{70.1} & 3.7 & 14.6 & \textbf{83.1} & 6.2 & \textbf{65.7} & \textbf{44.8} & 6.7 & 16.3\\
DSV4-Flash / DSH & 7 & 9/10 & \textbf{92.1} & \textbf{97.6} & \textbf{96.6} & \textbf{78.9} & \textbf{60.1} & \textbf{61.9} & \textbf{76.3} & \textbf{77.2} & \textbf{48.4} & 16.5\\
Qwen3.8-Max / OpenCode & 8 & 2/10 & 6.7 & 4.2 & \textbf{68.3} & 0.4 & \textbf{88.2} & 4.5 & 19.8 & 0.0 & 11.8 & 2.3\\
Qwen3.8-Max / Qwen Code & 9 & 5/10 & \textbf{40.8} & 16.0 & \textbf{46.8} & \textbf{63.7} & \textbf{69.8} & \textbf{75.0} & 0.5 & 18.6 & 0.0 & 0.0\\
Qwen3.7-Plus / OpenCode & 10 & 2/10 & 18.4 & 0.3 & 5.5 & 12.2 & 0.1 & 18.6 & \textbf{89.2} & 0.0 & \textbf{100.0} & 0.0\\
GLM-5.2 / OpenCode & 11 & 0/10 & 17.3 & 0.8 & 0.1 & 8.1 & 0.1 & 0.1 & 4.3 & 0.1 & 0.5 & 0.0\\
Kimi-K2.7-Code / OpenCode & 12 & 0/10 & 24.8 & 13.8 & 1.1 & 39.8 & 0.0 & 24.0 & 1.2 & 1.3 & 0.0 & 0.1\\
\bottomrule
\end{tabular}

\end{table}

\subsection{Performance--token frontiers}
\label{app:token-frontiers}
We repeat the frontier analysis using mean input-plus-output tokens per
appearance in place of dollars. Cached input and reasoning remain subsets
of input and output, respectively, and are counted once.
The cohort, Elo estimates, and 5,000 joint whole-match draws are unchanged.
This comparison removes price assumptions while retaining differences
in tokenization, context reuse, and model computation.
Native token counts do not represent equal compute across models.

\begin{figure}[!htbp]
\centering
\includegraphics[width=\linewidth]{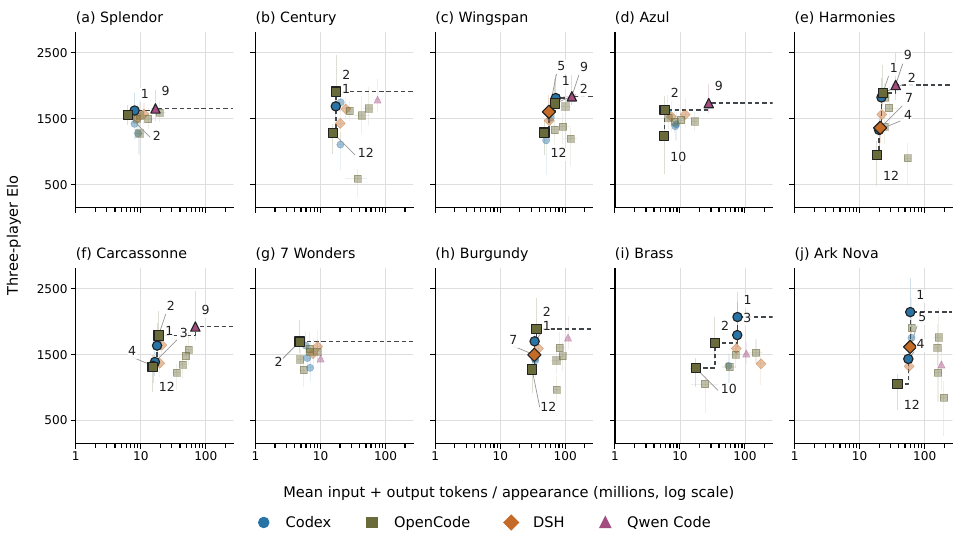}
\caption{Game-specific Elo--token Pareto frontiers on the same 480
three-player matches as Figure~\ref{fig:elo-cost}. Horizontal axes show
mean input-plus-output tokens per appearance in millions, on a log scale.
Elo axes, configuration IDs, and plotting conventions match the cost figure.
Outlined, labeled points are nondominated at the point estimates;
whiskers show pointwise 95\% whole-match intervals.
Token volume measures recorded usage, independently of the price schedule.}
\label{fig:elo-token}
\end{figure}

Token frontiers contain one to six configurations per game.
GPT-5.6-Sol/OpenCode appears in nine games under either resource measure.
GPT-5.6-Luna/Codex appears on ten dollar frontiers and three token frontiers;
DSV4-Pro/DSH appears on ten and two, respectively.
The differences reflect the price and token mix as well as recorded volume.
In 7 Wonders, GPT-5.6-Sol/OpenCode alone forms the token frontier:
it has both the lowest mean token volume and the highest estimated Elo.

\begin{table}[!htbp]
\centering\scriptsize
\setlength{\tabcolsep}{2.3pt}
\caption{Dollar and token frontier coverage, followed by token-frontier
membership frequencies (\%) over 5,000 joint whole-match draws.
USD and Tokens count point-estimate frontiers out of ten games.
Bold entries mark token-frontier membership at the point estimates.
Game abbreviations and IDs follow Table~\ref{tab:performance-compact}.}
\label{tab:token-frontier-stability}
\begin{tabular}{lrrr*{10}{r}}
\toprule
Configuration & ID & USD & Tokens & SP & CE & WI & AZ & HA & CA & 7W & BU & BR & AN\\
\midrule
GPT-5.6-Sol / Codex & 1 & 5 & 8 & \textbf{70.4} & \textbf{60.6} & \textbf{79.5} & 42.3 & \textbf{93.7} & \textbf{90.0} & 31.1 & \textbf{97.3} & \textbf{96.2} & \textbf{98.8}\\
GPT-5.6-Sol / OpenCode & 2 & 9 & 9 & \textbf{99.9} & \textbf{92.7} & \textbf{62.2} & \textbf{98.2} & \textbf{73.0} & \textbf{81.6} & \textbf{99.2} & \textbf{88.7} & \textbf{99.4} & 17.4\\
GPT-5.6-Terra / Codex & 3 & 3 & 2 & 7.7 & 11.9 & 31.2 & 0.4 & 67.9 & \textbf{59.9} & 1.1 & 26.1 & \textbf{49.5} & 16.5\\
GPT-5.6-Luna / Codex & 4 & 10 & 3 & 0.1 & 0.1 & 31.6 & 0.9 & \textbf{69.1} & \textbf{80.6} & 0.0 & 2.3 & 1.3 & \textbf{77.5}\\
DSV4-Pro / DSH & 5 & 10 & 2 & 15.9 & 1.2 & \textbf{92.3} & 16.9 & 31.0 & 5.7 & 16.7 & 1.8 & 24.0 & \textbf{47.3}\\
DSV4-Pro / OpenCode & 6 & 5 & 0 & 21.3 & 0.0 & 0.1 & 0.5 & 6.4 & 0.5 & 13.9 & 1.5 & 0.0 & 0.0\\
DSV4-Flash / DSH & 7 & 9 & 2 & 13.4 & 0.5 & 48.3 & 12.6 & \textbf{42.9} & 0.4 & 5.1 & \textbf{57.4} & 0.0 & 34.1\\
Qwen3.8-Max / OpenCode & 8 & 2 & 0 & 2.9 & 4.2 & 9.1 & 0.4 & 33.4 & 4.5 & 3.8 & 0.0 & 7.3 & 1.0\\
Qwen3.8-Max / Qwen Code & 9 & 5 & 5 & \textbf{43.0} & 16.0 & \textbf{47.1} & \textbf{63.7} & \textbf{70.1} & \textbf{75.0} & 0.5 & 18.6 & 0.0 & 0.0\\
Qwen3.7-Plus / OpenCode & 10 & 2 & 2 & 0.2 & 0.0 & 0.0 & \textbf{61.5} & 0.0 & 0.0 & 35.8 & 0.0 & \textbf{100.0} & 0.0\\
GLM-5.2 / OpenCode & 11 & 0 & 0 & 10.9 & 0.8 & 0.7 & 1.1 & 4.4 & 0.0 & 3.6 & 0.1 & 0.8 & 0.0\\
Kimi-K2.7-Code / OpenCode & 12 & 0 & 6 & 13.5 & \textbf{94.6} & \textbf{96.2} & 11.5 & \textbf{85.8} & \textbf{87.7} & 2.8 & \textbf{98.3} & 17.2 & \textbf{100.0}\\
\bottomrule
\end{tabular}

\end{table}

The cost and token analyses use identical outcomes, usage records, and
bootstrap draws. Cost-frontier membership frequencies agree across the
two implementations for all 120 configuration--game cells.
An independent sorted-usage sweep verifies all 120 token membership decisions.
Only the resource axis changes between the two analyses.
\FloatBarrier

\section{Supplementary Player-Count Results}
\label{app:player-count}
\subsection{Full native-system panel}
\label{app:player-count-panel}
The supplementary cohort contains 252 matches in five additional
game--player-count strata. Each configuration has 12 appearances per
stratum, balanced across seats: six per seat at two players and three at
four players. These results remain separate from the ten-game three-player
performance and resource analyses.

\begin{table}[!htbp]
\centering\footnotesize
\caption{Supplementary player-count PWS (\%) from 252 two-/four-player
matches, with 12 appearances per configuration and stratum. The common
three-player comparison across all ten games is in
Table~\ref{tab:performance-compact}. Only Splendor, 7 Wonders, and Brass have
additional observed player counts; coverage is listed in
Table~\ref{tab:expanded-exposure}.}
\label{tab:player-count-main}
\setlength{\tabcolsep}{3pt}
\begin{tabular*}{\linewidth}{@{\extracolsep{\fill}}lr*{5}{r}@{}}
\toprule
Configuration & ID & SP 2p & SP 4p & 7W 4p & BR 2p & BR 4p \\
\midrule
GPT-5.6-Sol / Codex & 1 & 50.0 & 55.6 & 77.8 & 100.0 & 91.7 \\
GPT-5.6-Sol / OpenCode & 2 & 58.3 & 52.8 & 58.3 & 83.3 & 66.7 \\
GPT-5.6-Terra / Codex & 3 & 58.3 & 29.2 & 56.9 & 91.7 & 77.8 \\
GPT-5.6-Luna / Codex & 4 & 16.7 & 13.9 & 36.1 & 29.2 & 44.4 \\
DSV4-Pro / DSH & 5 & 66.7 & 69.4 & 52.8 & 29.2 & 75.0 \\
DSV4-Pro / OpenCode & 6 & 58.3 & 47.2 & 62.5 & 50.0 & 38.9 \\
DSV4-Flash / DSH & 7 & 58.3 & 51.4 & 72.2 & 41.7 & 33.3 \\
Qwen3.8-Max / OpenCode & 8 & 75.0 & 63.9 & 55.6 & 33.3 & 33.3 \\
Qwen3.8-Max / Qwen Code & 9 & 41.7 & 80.6 & 41.7 & 54.2 & 63.9 \\
Qwen3.7-Plus / OpenCode & 10 & 8.3 & 33.3 & 16.7 & 16.7 & 30.6 \\
GLM-5.2 / OpenCode & 11 & 50.0 & 61.1 & 27.8 & 20.8 & 36.1 \\
Kimi-K2.7-Code / OpenCode & 12 & 58.3 & 41.7 & 41.7 & 50.0 & 8.3 \\
\bottomrule
\end{tabular*}

\end{table}

Table~\ref{tab:player-count-main} adds the observed two-/four-player strata
to the ten-game three-player matrix in Table~\ref{tab:performance-compact}.
In Splendor, Qwen3.8-Max / OpenCode leads the two-player PWS estimates (75.0\%),
whereas Qwen3.8-Max / Qwen Code leads at three and four players (70.8\% and
80.6\%). Across all 12 configurations, the two--three-player PWS-rank
correlation is 0.070 in Splendor versus 0.701 in Brass; the three--four-player
correlations are 0.792, 0.552, and 0.880 in Splendor, 7 Wonders, and Brass.
Player-count mixtures can change the observed ordering within a game.

There is no uniform benefit from a larger table. DSV4-Pro / DSH's Brass PWS
rises from 29.2\% to 75.0\% between two and four players, while GPT-5.6-Sol / Codex
moves from 100\% to 91.7\%. These are descriptive schedule-conditioned
contrasts, since opponent composition also changes.
These results motivate the fixed three-player main comparison.

\section{Matched Harness Studies}
\label{app:bridge}

\subsection{Design, evidence, and estimands}

The bridge crosses DSV4-Flash
and Qwen3.8-Flash with two harnesses in two games, nine scenarios per game. Each scenario
specifies one setup seed and one focal seat and contains four separate
three-player matches. The other seats use DSV4-Flash / OpenCode and
Qwen3.8-Flash / OpenCode as fixed opponents. The four focal configurations do
not play one another in a single match. Comparisons therefore describe
performance against this anchor population, not a four-participant league.
The bridge uses the DeepSeek V4 Flash 0731 snapshot.
Each focal seat occurs three times in every model--harness--game cell.
The listed schedule places OpenCode first in 18 of the 36 harness pairs
and Qwen Code first in 18.

The analysis includes 72 matches arranged into 36 matched harness pairs.
We verify that each scheduled match has one result, with matching seeds,
seats, and opponent configurations within each pair. We also check all paired
score differences and aggregate point estimates. Every focal rank is
1, 2, or 3 and agrees with PWS $Y=(3-\mathrm{rank})/2$ and the first-place
indicator. These checks establish consistency between the schedule, focal
outcomes, and summary statistics. Execution records and participant
settings document the evaluated conditions
(Appendix~\ref{app:reproducibility}).

For scenario $b$ in game $g$, let $Y_{m,h,g,b}$ be final focal PWS.
The observed harness contrast and model-dependent
contrast are
\begin{align}
\widehat\Delta_{m,g}
&=\frac{1}{9}\sum_{b=1}^{9}
  \left(Y_{m,\text{Qwen Code},g,b}-Y_{m,\text{OpenCode},g,b}\right),\\
\widehat I_g
&=\widehat\Delta_{\text{Qwen3.8-Flash},g}
  -\widehat\Delta_{\text{DSV4-Flash},g}.
\label{eq:bridge-interaction}
\end{align}
Overall contrasts average the two
games equally. No raw game scores are pooled across titles. These
estimands concern final competitive performance. Shared seeds do not imply
shared subsequent trajectories:
the focal changes can alter both opponents' decisions and future states.

\subsection{Uncertainty and observed ordering}

We draw 100,000 bootstrap replicates with RNG seed 20260917. Within each
game, nine scenarios are sampled with replacement, retaining all four
model--harness outcomes from every selected scenario. This preserves the
dependence needed for model comparisons and differences in harness gains.
The two game means are averaged for pooled replicates. Intervals are
pointwise empirical percentiles, conditional on the selected games and
anchor configurations.

\begin{table}[t]
\centering\small
\caption{Focal PWS (\%) and harness contrasts (percentage points), Qwen Code minus OpenCode. Each game--model cell contains nine matched pairs; the pooled row has 18 pairs with equal game weights. Pointwise 95\% percentile intervals use 100,000 scenario-cluster bootstrap draws. These are anchor-relative PWS diagnostics, not Elo or four-way head-to-head rankings.}
\label{tab:bridge-pws}
\begin{tabular}{llrrr}
\toprule
Game & Focal model & OpenCode & Qwen Code & $\Delta$ PWS [95\% CI] \\
\midrule
Splendor & DSV4-Flash & 50.0 & 55.6 & $+5.6\;[-38.9,44.4]$ \\
Splendor & Qwen3.8-Flash & 38.9 & 55.6 & $+16.7\;[-16.7,50.0]$ \\
7 Wonders & DSV4-Flash & 66.7 & 61.1 & $-5.6\;[-38.9,27.8]$ \\
7 Wonders & Qwen3.8-Flash & 27.8 & 72.2 & $+44.4\;[11.1,77.8]$ \\
Equal-game mean & DSV4-Flash & 58.3 & 58.3 & $+0.0\;[-27.8,25.0]$ \\
Equal-game mean & Qwen3.8-Flash & 33.3 & 63.9 & $+30.6\;[5.6,52.8]$ \\
\bottomrule
\end{tabular}

\end{table}

The point-estimate model ordering changes across harnesses, but the
pooled model--harness contrast remains uncertain: $\widehat I=+30.6$ pp,
95\% interval $[-5.6,69.4]$. Within Splendor it is $+11.1$
$[-44.4,72.2]$ pp; within 7 Wonders it is $+50.0$ $[11.1,100.0]$ pp.
The difference between these game-specific contrasts is $+38.9$ pp
$[-38.9,116.7]$, not a resolved three-way interaction.
As a sensitivity check for the family of four model--game harness
contrasts, 98.75\% percentile intervals are used at the nominal
Bonferroni level. Qwen3.8-Flash's 7 Wonders interval becomes $[0.0,83.3]$ pp;
all four include zero.

\begin{table}[t]
\centering\small
\caption{Anchor-relative final placements in the matched bridge. Each row
has 18 matches, nine per game. Rows follow point-estimate PWS.
Equal PWS can conceal different
first-place and last-place frequencies.}
\label{tab:bridge-ranks}
\begin{tabular}{llrrrr}
\toprule
Focal model & Harness & PWS (\%) & First & Second & Third\\
\midrule
Qwen3.8-Flash & Qwen Code & 63.9 & 9 & 5 & 4\\
DSV4-Flash & OpenCode & 58.3 & 7 & 7 & 4\\
DSV4-Flash & Qwen Code & 58.3 & 6 & 9 & 3\\
Qwen3.8-Flash & OpenCode & 33.3 & 3 & 6 & 9\\
\bottomrule
\end{tabular}

\end{table}

\subsection{48-match within-Qwen-family study}
\label{app:qwen-harness-pilot}

\paragraph{Design and scope.}
The independent study contains 48 complete three-player matches:
Qwen3.8-Flash and Qwen3.8-Max, OpenCode and Qwen Code, Splendor and 7 Wonders,
and six scenarios per game. Each scenario fixes the seed and focal seat
across four configurations, using each seat twice per model--harness--game
cell. Anchors are Qwen3.8-Flash / OpenCode and Qwen3.8-Max / OpenCode,
with separate seat-scoped sessions. The frozen schedule randomizes scenario
and configuration order. Matches start with independent workspaces and
native sessions, a 1,200-second external-process limit per decision, and
at most 32 concurrent matches. Registered model identifiers and effort
settings are constant across harnesses.

\paragraph{Paired outcomes.}
Let $Y_{m,h,g,b}$ be the focal player's PWS in a complete match.
For this study, the harness contrast is
$\widehat\Delta_{m,g}=\frac{1}{6}\sum_{b=1}^{6}
(Y_{m,\text{Qwen Code},g,b}-Y_{m,\text{OpenCode},g,b})$.
The model contrast is
$\widehat I_g=\widehat\Delta_{\text{Qwen3.8-Max},g}
-\widehat\Delta_{\text{Qwen3.8-Flash},g}$.
This direction differs from the cross-model-family contrast only in its model
panel. Different anchors keep the studies separate despite their shared
Qwen3.8-Flash identifier. Neither contributes to native Elo fits.

\begin{table}[!htbp]
\centering\small
\caption{The 48-match within-Qwen-family study. Each model--game comparison
has six matched pairs. PWS is in percent and contrasts are in percentage
points; $+/=/-$ counts scenarios in which Qwen Code improves, matches,
or worsens focal placement relative to OpenCode. Intervals are pointwise
95\% scenario-bootstrap percentiles.}
\label{tab:qwen-pilot-pws}
\setlength{\tabcolsep}{4pt}
\begin{tabular}{llrrrr}
\toprule
Game & Model & OpenCode & Qwen Code & $\Delta$ [95\% interval] & $+ / = / -$\\
\midrule
Splendor & Qwen3.8-Flash & 25.0 & 33.3 & $+8.3\;[-25.0,41.7]$ & 3/1/2\\
Splendor & Qwen3.8-Max & 33.3 & 91.7 & $+58.3\;[50.0,75.0]$ & 6/0/0\\
7 Wonders & Qwen3.8-Flash & 16.7 & 33.3 & $+16.7\;[-25.0,58.3]$ & 3/1/2\\
7 Wonders & Qwen3.8-Max & 75.0 & 83.3 & $+8.3\;[-25.0,50.0]$ & 1/4/1\\
\bottomrule
\end{tabular}

\end{table}

\paragraph{Uncertainty and multiplicity.}
We enumerate all $6^6=46{,}656$ ordered bootstrap resamples within each
game, retaining the four configurations together in every selected scenario.
Figure~\ref{fig:matched-bridge}b--d and Table~\ref{tab:qwen-pilot-pws}
show the resulting paired estimates. All six Qwen3.8-Max/Splendor pairs
improve, by 100 pp in one scenario
and 50 pp in the other five. The two-sided sign-test value is
$p=0.03125$ before adjustment and $0.125$ under a Bonferroni correction
for the four model--game harness contrasts. The corresponding Qwen3.8-Max-minus-Qwen3.8-Flash
gain is +50.0 pp [16.7, 83.3], but only four scenarios have nonzero
interaction differences; their two-sided sign test gives $p=0.125$.
In 7 Wonders, the interaction is $-8.3$ pp [$-83.3$, 66.7].

\paragraph{Paired placements.}
Qwen3.8-Max's first/second/third counts in Splendor are 5/1/0 under Qwen Code
versus 0/4/2 under OpenCode; in 7 Wonders it wins four of six under either
harness. Qwen3.8-Flash's paired changes are mixed in both games
(Figure~\ref{fig:matched-bridge}c).

\paragraph{Outcome verification.}
The outcome ledger contains 48 matches and 24 harness pairs. Exact rational
arithmetic reconciles every PWS with the terminal ranks. All 48 final game records
pass integrity checks and match the stored identity, rules version, seed,
terminal status, and turn count. Paired statistics,
Table~\ref{tab:qwen-pilot-pws}, and Figure~\ref{fig:matched-bridge} are computed
directly from these outcomes. Checksums verify source-record integrity.

\end{document}